\documentclass[12pt,arial,onecolumn,letterpaper]{elsarticle}

\usepackage[margin=0.8in]{geometry}
\usepackage{url}
\usepackage[margin=0.8in]{geometry}
\usepackage{url}
\usepackage{xcolor}
\usepackage{amsmath}
\usepackage{amssymb}
\usepackage{xspace}
\usepackage{booktabs}
\usepackage{enumerate}
\usepackage{mathtools}

\usepackage{amssymb}

\usepackage{times}
\usepackage{epsfig}
\usepackage{graphicx}
\usepackage{amsthm}
\usepackage{amsmath}
\usepackage{amssymb}
\usepackage{booktabs}
\usepackage{subfigure}
\usepackage{algorithm,algorithmic}
\usepackage{color}
\usepackage{caption}
\usepackage{multirow}
\usepackage{array}
\usepackage{pdfpages}
\usepackage{siunitx}



\def\eg{\textit{e.g.,~}}
\def\ie{\textit{i.e.,~}}
\def\etal{\textit{et al.}~}
\def\Vec#1{{\boldsymbol{#1}}}
\def\Mat#1{{\boldsymbol{#1}}}
\newtheorem{theorem}{Theorem}[section]

\newtheorem{proposition}[theorem]{Proposition}

\begin{document}

\begin{frontmatter}



\title{\textbf{What is the Best Way for Extracting Meaningful Attributes \\ from Pictures?}}


\author{
Liangchen Liu, Arnold Wiliem, Shaokang Chen, Brian C. Lovell \\
The University of Queensland, School of ITEE QLD 4072, Australia \\
{\footnotesize	 \tt l.liu9@uq.edu.au,a.wiliem@uq.edu.au,shaokangchenuq@gmail.com,lovell@itee.uq.edu.au} \\
project page: http://comingsoon}
\begin{abstract}
\label{abstract}
Automatic attribute discovery methods have gained in popularity to extract sets of visual attributes from images or videos for various tasks.
Despite their good performance in some classification tasks, it is difficult to evaluate whether the attributes discovered by these methods are meaningful and which methods are the most appropriate to discover attributes for visual descriptions. 
In its simplest form, such an evaluation can be performed by manually verifying whether there is any consistent identifiable visual concept distinguishing between positive and negative exemplars labelled by an attribute.
This manual checking is tedious, expensive and labour intensive.
In addition, comparisons between different methods could also be problematic as it is not clear how one could quantitatively decide which attribute is more meaningful than the others. 
In this paper, we propose a novel attribute meaningfulness metric to address this challenging problem.
With this metric, automatic quantitative evaluation can be performed on the attribute sets; thus, reducing the enormous effort to perform manual evaluation. 
The proposed metric is applied to some recent automatic attribute discovery and hashing methods on four attribute-labelled datasets. 
To further validate the efficacy of the proposed method, we conducted a user study. 
In addition, we also compared our metric with a semi-supervised attribute discover method using the mixture of probabilistic PCA.
In our evaluation, we gleaned several insights that could be beneficial in developing new automatic
attribute discovery methods. 
\end{abstract}


\end{frontmatter}



\section{Introduction}
\label{sec_introduction}

\begin{figure}
\centering
    \includegraphics[width=0.7\linewidth]{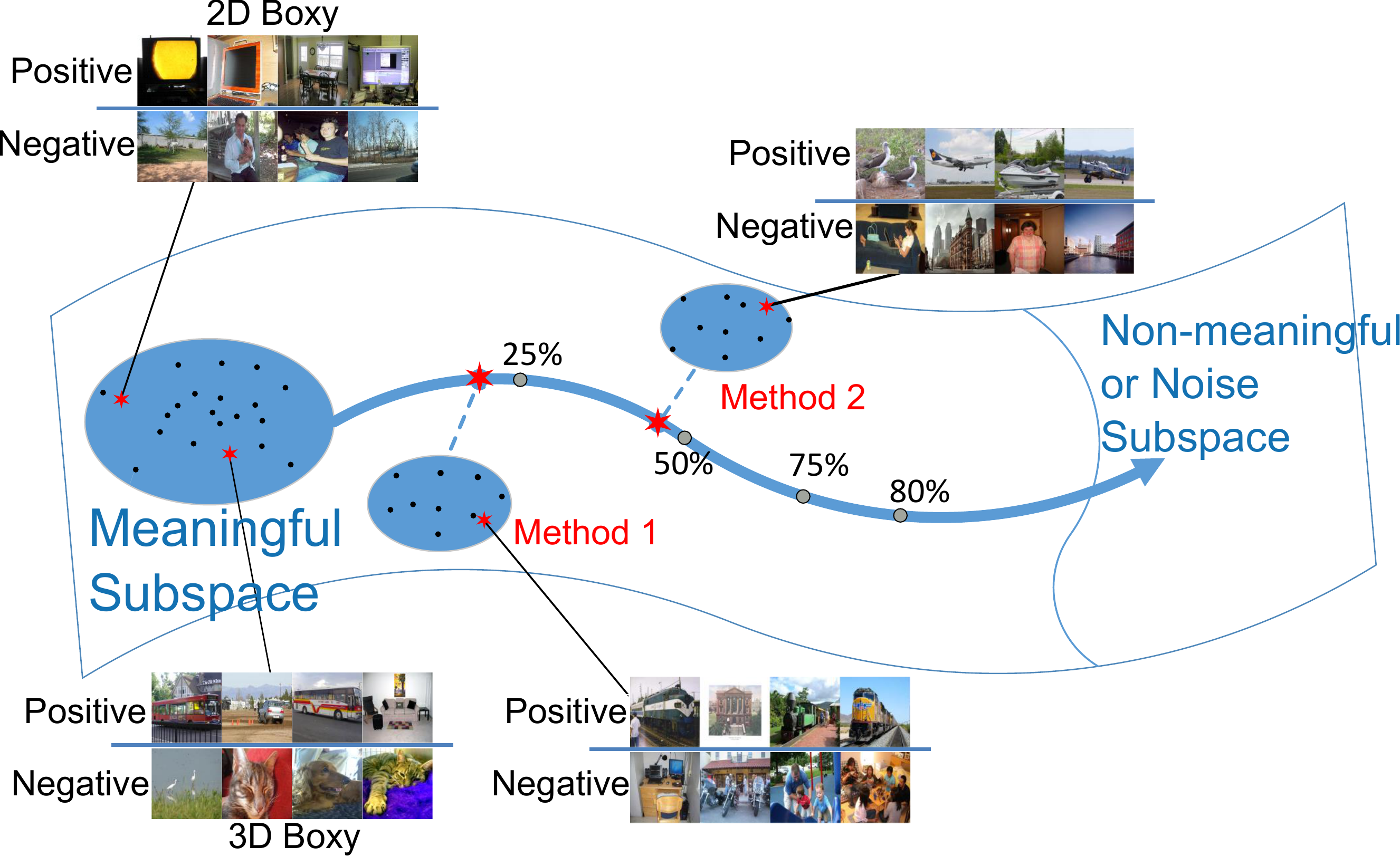}
    \caption{An illustration of the proposed attribute meaningfulness metric.
    Each individual attribute is represented as the outcome of the corresponding attribute classifier tested on a set of images.
    Inspired by~\cite{parikh2011interactively} we propose an approach to measure the distance between a set of discovered attributes and the meaningful subspace.
     The metric score is derived using a subspace interpolation between Meaningful Subspace and Non-Meaningful/Noise Subspace.
     The score indicates how many meaningful attributes are contained in the set of discovered attributes.
    }
\label{fig:idea}
\end{figure}

Language is one of the most important factors in communication.
We would not have been able to write this paper if there was not any language!  
Human language has been used for solving computer vision problems such as scene understanding~\cite{Patterson2012SunAttributes} and image or video description\cite{liu2013video,rohrbach14gcpr} and image retrieval~\cite{liu2007survey}. 
The language model helps us to make an effective transfer of domain expert knowledge into machines.
People often say that ``a picture is worth a thousand words.'' Turning this around we can also say that ``a thousand words/visual attributes are required to explain a picture.'' 
The latter form can be quite powerful to address many computer vision problems~\cite{Farhadi09describingobjects,kumar2009attribute,feng2014learning,ChangYLZH16,macomplex}.
For instance the active learning framework proposed in~\cite{Kovashka2015} employs human knowledge to learn better models.
Visual attributes are extremely useful as they are: (1) human understandable; (2) machine computable and (3) shared across classes.
For these reasons, recently many attribute discovery methods have been proposed to extract visual/image attributes~\cite{bergamo2011picodes,rastegari2012attribute,kovashka2015discovering}.
Broadly, several more concrete visual attributes such as color and texture have attracted great attentions~\cite{zhang2016spatiochromatic,dong2015texture}.
To that end, color saliency analysis~\cite{zhang2016spatiochromatic} and shearlets-based texture learning~\cite{dong2015texture} are proposed and have achieved promising performance.

One of the biggest challenges in using attribute descriptors
is that an enormous amount of training images with attribute labels is required to train attribute classifiers. 
It is extremely tedious, time-consuming and expensive to label each individual image for every attribute (\eg if there are 64 attributes, then each image should have at least 64 attribute labels).
Furthermore, in some specialized domains such as \textit{Ornithology}~\cite{WelinderEtal2010}, \textit{Entomology}~\cite{Wang2009learning} and cell pathology~\cite{Wiliem2014discovering}, the human labelling task could be immensely expensive as only a few highly trained experts could conduct such a task.

To reduce the workload, automatic attribute discovery methods have been developed~\cite{bergamo2011picodes,rastegari2012attribute,sharmanska2012augmented,Wiliem2014discovering,yfx2013cvpr,zhang2012review}. 
The primary aim of these works is to learn a function that maps the original image feature space into a binary 
code space wherein each individual bit represents the presence/absence of a visual attribute.
These attribute discovery methods are also closely related to hashing methods~\cite{gong2011iterative,Leskovec2009Mining,weiss2009spectral}.
The difference is that unlike automatic attribute discovery approaches,
hashing methods focus more on how to significantly reduce the storage demand and computational complexity whilst maintaining system accuracy.

Despite great strides that have been made in this field, there are still some important open questions: left unaddressed: 
1) Given the set of attributes/binary codes discovered by a method, are these attributes or binary codes really meaningful? 
2) Can we compare these methods by directly observing the discovered attributes?
By exploring these questions, we can begin to glean some insights on mechanisms required to extract meaningful attributes/binary codes. 
We note that the aim of this work is not to propose a new method to discover attributes. Instead, we 
propose a novel meaningfulness metric and use this tool to study the existing methods. 

Gauging ``how meaningful'' for a given attribute can be an ill-posed problem as there is no obvious \textit{yardstick} for measuring this. 
Fortunately, it is pointed out by Parikh and Grauman that meaningful attributes may have a \textit{shared structure}~\cite{parikh2011interactive,parikh2011interactively}.
This means, given the attribute feature space, meaningful attributes are likely to be close to each other within a subspace.
In~\cite{Liu2016isba}, we further studied this shared structure and applied our findings to the task of automatic generation of surveillance video descriptions.

Inspired by these works, we propose a novel meaningfulness metric that could become one of the \textit{yardsticks} to measure attribute meaningfulness. 
More specifically, 
we first measure the distance between the discovered attributes and the meaningful attribute subspace. 
To this end, an approximate geodesic distance based on reconstruction error is proposed. 
As it may be difficult to perform quantitatively analysis/study using this distance directly, we then derive the meaningfulness metric based on the distance. 
In particular, the metric is derived by first performing a subspace interpolation between meaningful subspace and non-meaningful subspace lying on the manifold of decision boundaries. 
The distance on each interpolated subspace is calculated.
These are then used to calibrate the distance of the discovered attributes to the meaningful subspace. 
Fig.~\ref{fig:idea} illustrates our main idea.

An earlier form of this work has been presented in [20]. 
In this work, we extend our earlier work in several aspects.
We perform in-depth analysis on the proposed metric and consider additional dataset.
Throughout these experiments we found that the calibration step heavily depends on the meaningful subspace spanned by the selected meaningful attributes, denoted the subspace bases.
The space spanned by these bases should be maximized in order to ensure the calibration is done properly.
To remedy this, we present a simple-yet-effective technique using semantic reasoning and threshold setting.

\noindent
\textbf{Contributions --- } We list our contributions as follows:
\begin{itemize}
	\vspace{-1.5ex}
    \item  We propose a reconstruction error based approach with two different regularizations (\ie $\ell_0$ and convex hull) to approximate the geodesic distance between a given attribute set and the meaningful subspace.
	\vspace{-1.5ex}
    \item We propose the novel \textit{attribute meaningfulness} metric that
    allows us to quantitatively measure the meaningfulness of a set of automatically discovered attributes.
    The metric score is related to ``the percentage of meaningful attributes contained in the set of attributes.''
    \vspace{-1.5ex}
    \item We propose an improved calibration method to avoid pathological cases where the calibration could not be performed. 
    This method is developed based on the in-depth analysis performed in this work.
    \vspace{-1.5ex}
    \item We present extensive experiments and analysis on four popular attribute-labelled datasets to demonstrate that our proposal can really capture attribute meaningfulness.
     The \textit{attribute meaningfulness} of some recent automatic attribute discovery methods and various hashing approaches are also evaluated on these datasets.
    A user study is conducted to further validate the effectiveness of the proposed metric. 
    In addition, we compare the proposed metric with a metric adapted from a recent semi-supervised attribute discovery method using the Mixture of Probabilistic Principal Component Analysis (MPPCA)~\cite{parikh2011interactively,tipping1999mixtures}. 
\end{itemize}

\vspace{-1ex}
We continue our paper as follows.
Related work is discussed in Section~\ref{sec_related}.
Then we introduce our approach of evaluating attribute meaningfulness in Section~\ref{sec:measuring_meaningfulness}.
Our proposed metric is described in Section~\ref{sec:part3}.
Next we discuss the experiments and results in Section~\ref{Experiment}.
Finally the main findings and future directions are presented in Section~\ref{sec_conclusion}.

\section{Related Work}
\label{sec_related}
Evaluation of visual attribute meaningfulness is traditionally conducted by manually checking the presence/absence of consistent identifiable visual concepts in a set of given images.
This task usually requires a large-scale human labelling effort.
A system such as the Amazon Mechanical Turk~(AMT)~\footnote{www.mturk.com} is able to handle this task for small datasets.
However, since this process needs to be repeated whenever new attributes are discovered or novel methods are proposed, this manual process is ineffective and expensive.
In our case, the AMT Human Intelligence Task (HIT) is to evaluate the meaningfulness of attributes by examining corresponding positive and negative images according to each attribute.
The average time of each worker spent on this typical HIT is 2 minutes~\cite{Patterson2012SunAttributes}.
Then an AMT worker may require 320 minutes to evaluate 32
attributes discovered by 5 different methods (\ie $32 \times 5
\times 2 = 320$ minutes). 
The time spent could increase significantly if statistically reliable results are desired by increasing the number of  AMT workers.

Unequivocally, it is more desirable to develop an automatic approach, which is more cost-effective, less labor intensive and time consuming to evaluate the meaningfulness of the set of discovered attributes.
The task of measuring the \textit{attribute meaningfulness} of discovered attributes is similar to the task presented in the Turing Test \cite{turing1950computing}. 
In this task, we would like to measure how much a machine could provide responses like a human being. 
If a machine could respond like a human being, it means that the results produced must have significant meaning. 
Unfortunately, the Turing Test still requires a human judge who actively engages with the machine.

To that end, several works~\cite{baird2003pessimalprint, Rui2004Automated, turing1950computing} aim to devise an automated Turing Test that follows the framework of this
famous test but replaces the human judge by another machine. 
A notable example is CAPTCHA~\cite{von2003captcha} which is very prevalent in web security applications. 
This technique basically lets a machine be the judge issuing the test to determine whether the subject is a human. 
Generally, CAPTCHA provides a challenge in the form of an image containing numbers or characters which are difficult to be identified
by current machines. 
The main assumption in CAPTCHA is that machine recognition will not be as good as human.

Our work can be interpreted as an instance of the automated Turing Test as follows.
We are testing a set of automatic attribute discovery techniques by giving challenges in the form of images. 
These techniques are then giving us a set of attributes.
We will automatically verify the meaningfulness through the positive and negative images generated from each attribute classifier.
Note that if we have human observers performing the verification instead of machines, then this becomes an instance of the standard Turing Test.
To perform the automated Turing Test, there has to be a measurement to determine which automatic attribute discovery technique is `good' and which one is `not that good'.

Some unsupervised semantic visual representation learning works~\cite{huang2016unsupervised,hong2016joint,rastegari2012attribute,Wiliem2014discovering} have indicated that it is possible to discover the meaningful concepts unsupervisedly from data itself with or without side information.
Such as Chen~\etal~\cite{huang2016unsupervised} introduce a simple yet powerful unsupervised approach to learn and predict visual attributes directly from data.
With the help of deep Convolutional Neural Networks (CNNs), they train to output a set of discriminative, binary attributes often with semantic meanings.
Hong~\etal~\cite{hong2016joint} propose a novel algorithm to cluster and annotate a set of input images with semantic concepts jointly.
They employ non-negative matrix factorization with sparsity and orthogonality constraints to learn the label-based representations with the side information (a labeled reference image database) obtaining promising results.

All of these works imply there may be some potential relations between meaningful concepts.
Fortunately, the shared structure assumption among meaningful attributes proposed in~\cite{parikh2011interactively} can serve as the foundation of the automatic measurement.
Based on this assumption, Parikh and Grauman \etal proposed an active learning approach that uses Mixtures of Probabilistic Principal Component Analysers (MPPCA)~\cite{tipping1999mixtures}
to predict how likely an attribute is nameable. 
Nevertheless, their work only focuses on deciding whether an attribute is nameable or not.
Their work does not tackle the problem of quantitatively measuring the attribute meaningfulness. 
In addition, this approach requires human interaction to populate the nameability space.
Thus, their method is not suitable for addressing our goal 
(i.e., to automatically evaluate the meaningfulness of attribute sets).

In our previous work, the shared structure assumption is utilized~\cite{Liu2016isba}.
In particular, the work in~\cite{Liu2016isba} proposed a selection approach of attribute discovery methods to assist attribute-based keywords generation for video description from surveillance systems.
However, the work did not consider quantitative analysis of the meaningfulness of the discovered attributes (\eg how much meaningful content is contained in a set of automatically discovered attributes). 
In addition, the characteristics of the meaningfulness of attributes may vary to some extent. 


\section{Measuring Attribute Set Meaningfulness}
\label{sec:measuring_meaningfulness}




We begin by describing the manifold of decision boundaries and the meaningful attribute subspace. 
Then, we define the distance between the automatically discovered attributes and the meaningful attribute set in the manifold space to measure the attribute meaningfulness.

\subsection{Manifold of decision boundaries}
\label{sec:manifold}

Supposed there is a set of exemplars $\mathcal{X} = \{ \Vec{x}_i \}_{i=1}^N$, an attribute can be considered as a decision boundary which partitions the set into two subsets $\mathcal{X}^+ \cup \mathcal{X}^- = \mathcal{X}$. Here $\mathcal{X}^{+}$ represents the set where the attribute is present and $\mathcal{X}^{-}$ represents the set where the attribute is absent.
Therefore, all the attributes are lying on a manifold formed by decision boundaries~\cite{parikh2011interactively}.

In this case, an attribute can also be viewed as an $N$-dimensional binary vector whose element represents the classification output of all exemplars $\Vec{x}_i$ classified by the corresponding attribute binary classifier $\phi ( \cdot ) \in \mathbb{R}$.
The sign of the classifier output on $\Vec{x}_i$ indicates whether the exemplar belongs to the positive or negative set (\ie $\mathcal{X}^+$ or $\mathcal{X}^-$). 
As such, an attribute can be represented as $\Vec{z}^{[\mathcal{X}]} \in \{-1,+1\}^{N}$ whose $i$-th element is $\Vec{z}_{(i)}^{[\mathcal{X}]} = \operatorname{sign}(\phi(\Vec{x}_i)) \in \{-1,+1\}$.
For the sake of simplicity, we drop the symbol $[\mathcal{X}]$ from $\Vec{z}^{[\mathcal{X}]}$ whenever the context is clear.
Thus, the manifold of decision boundaries w.r.t. $\mathcal{X}$ can be defined as 
$\mathcal{M}^{[\mathcal{X}]} \in \{-1,+1\}^{N}$ which is embedded in a $N$-dimensional binary space.
Again, we also write $\mathcal{M}^{[\mathcal{X}]}$ as $\mathcal{M}$.

As observed from~\cite{parikh2011interactively,parikh2011interactive}, the meaningful attributes have shared structure wherein they lie close to each other on the manifold. In other words all the meaningful attributes are contained in a subspace on $\mathcal{M}$. 
In an ideal case, all possible meaningful attributes should be enumerated in this subspace. 
Unfortunately, it is infeasible to enumerate all of them. 
One intuitive solution is by relying on the existing human knowledge, that is the human labelled attributes from various datasets such as \cite{biswas2013simultaneous,parikh2011interactive,parikh2011interactively}. 
These attributes are all naturally meaningful since they are collected via manual human labeling process using the AMT. 
However, the number of available labelled attributes may not be sufficient. 
To this end, based on the shared structure assumption, we thus introduce an approximation of the meaningful subspace by linear combinations of the human labelled attributes. 
This means, if an automatically discovered attribute is close enough to any attribute existing in the meaningful subspace, it should be considered as a meaningful attribute. 

\subsection{Distance of an attribute to the Meaningful Subspace}
\label{sec:distance}
In this section, we mathematically describe the reconstruction error based distance of an attribute to the Meaningful Subspace.
Given a set of $N$ images $\mathcal{X}$, we denote $\mathcal{S} = \{\Vec{h}_j\}_{j=1}^J, \Vec{h}_j \in \{-1,+1\}^N$ as the set of meaningful attributes.
We use a matrix $\Mat{A} \in \mathbb{R}^{N \times J}$, in which each column vector is the representation of an attribute, to form the set $\mathcal{S}$.
As the assumption in~\cite{parikh2011interactively}, meaningful attributes should be close to the meaningful subspace spanned by the set of meaningful attributes $\mathcal{S}$.
For instance, the primary colors \ie red, green, blue are able to construct the set of secondary colors such as yellow, magenta and cyan.
Moreover, the primary colors can provide negative information clues to describe other primary colors ~(\eg blue is neither green nor red).
Under this assumption, we are able to define a reconstruction error based distance between an attribute and the meaningful subspace.
More specifically, let $\Vec{z}_k$ be an attribute and $\Mat{A}$ be the representation of meaningful attributes. The distance is defined as:
\begin{equation}
	\min_\Vec{r} \| \Mat{A} \Vec{r}  - \Vec{z}_k \|_2^2,
	\label{eq:single_dist}
\end{equation}
\noindent
where $\Vec{r} \in \mathbb{R}^{J \times 1}$ is  the reconstruction coefficient vector.
Note that the reconstruction in \eqref{eq:single_dist} may not be in the manifold $\mathcal{M}$ (i.e. $\Mat{A}\Vec{r} \notin \mathcal{M}$). 
Therefore, we relax this reconstruction procedure into Euclidean space for computational simplicity.
This relaxation effectively becomes an approximation of the true geodesic distance.

\subsection{Distance between a set of discovered attributes and the Meaningful Subspace}
Analogously, suppose there are $K$ discovered attributes, we use matrix $\Mat{B} \in \{0, 1\}^{N \times K}$ to represent the discovered attribute set $\mathcal{D}$.
Then, according to the specific set of images $\mathcal{X}$, we can define the distance between the set of discovered attributes $\mathcal{D}$ and the Meaningful Subspace $\mathcal{S}$ as the average reconstruction error:
\begin{equation}
	\delta(\mathcal{D},{\mathcal{S}}; \mathcal{X}) = \frac{1}{K}\min_{\Mat{R}} \| \Mat{A}\Mat{R} - \Mat{B} \|_F^2,
	\label{eq:min_dist}
\end{equation}
where $\| \cdot \|_F$ and $\Mat{R} \in \mathbb{R}^{J \times K}$ are the Frobenious norm and the reconstruction matrix respectively.

The reconstruction coefficients are preferably sparse, because generally only a few attributes can provide useful clues to reconstruct a particular attribute while most of them should stay inactive in this procedure. 
Similar to the example in section~\ref{sec:distance}, only a few color attributes can reconstruct another color attribute, most of them should stay inactive (\ie their reconstruction coefficient should be 0).
Unfortunately, the distances in~\eqref{eq:single_dist} and~\eqref{eq:min_dist} may create dense 
reconstruction coefficients due to the absence of a regularization term.
As such, we first introduce convex hull regularization used in~\cite{Liu2016isba}.
Moreover, according to~\cite{serre2007robust}, the perception mechanism of human visual systems follows the sparsity principle.
That means only a few attributes will first trigger the semantic-visual connection in our brain.
Desirable attribute discovery methods should also obey this principle. 
Hereby, we consider the sparsity-inducing $\ell_0$ regularization as the second regularization alternative.

\vspace{-1ex}
\subsubsection{Convex hull regularization}

Via introducing a convex hull constraint,~\eqref{eq:min_dist} becomes:
\begin{equation}
\label{eq:con_hull}
\begin{aligned}
	\delta_{\operatorname{cvx}}(\mathcal{D},\mathcal{S}; \mathcal{X}) = \frac{1}{K}\min_{\Mat{R}} \| \Mat{A}\Mat{R} - \Mat{B} \|_F^2   \operatorname{s.t.} \\
     {\Mat{R}}\left( {i,j} \right) \ge 0 & \\
     \sum_{i=1}^{J} {\Mat{R}(i,\cdot)}=1. &
\end{aligned}	
\end{equation}
\noindent
This objective function describes the average distance between each discovered attribute $\Vec{z}_k \in \mathcal{D}$ and the convex hull of $\mathcal{S}$. 
Its optimization can be efficiently solved using the method proposed in~\cite{cevikalp2010face}.

\subsubsection{$\ell_0$ regularization}
As to $\ell_0$ regularization, different from the convex hull regularization, a possible 
direct correlation between each discovered attribute $\Vec{z}_k \in \mathcal{D}$ 
and the meaningful attribute, $\Vec{h}_j \in \mathcal{S}$ is considered:

\begin{align}	
	\label{eq:joint_prob}
		\delta_{\operatorname{jp}}(\mathcal{D},{\mathcal{S}}; \mathcal{X}) = & \frac{1}{K} \min_{\Mat{R}} \| \Mat{A}\Mat{R} - \Mat{B} \|_F^2, \operatorname{s.t.} \\
		& \forall k \in \{ 1 \cdots K \}, \| \Mat{R}_{\cdot,k} \|_0 \leq 1, \nonumber \\
		& \forall j \in \{ 1 \cdots J \}, \| \Mat{R}_{j,\cdot} \|_0 \leq 1. \nonumber	
\end{align}			
\noindent
where $\Mat{R}_{j,\cdot}$, $\Mat{R}_{\cdot,k}$ represent the $j$-th row vector and the $k$-th column vector in matrix $\Mat{R}$ respectively.
The two additional $\ell_0$ regularizers enforce one-to-one relationships between ${\mathcal{S}}$ and $\mathcal{D}$.
The reconstruction matrix $\Mat{R}$ correlates each discovered attribute to a particular meaningful attribute.
More specifically, for each discovered attribute $\Vec{z}_k \in \mathcal{D}$,
the closest $\Vec{h}_j \in \mathcal{S}$ is found to minimize the function.
However, it could be possible that $|{\mathcal{S}}| > |\mathcal{D}|$.
In this case, we can only match $K$ discovered attributes in ${\mathcal{S}}$ and vice versa.

Unfortunately, the optimization for~\eqref{eq:joint_prob} is non-convex.
As such, a greedy approach is proposed to address this
through iteratively finding pairs of meaningful discovered attributes with the smallest distance.
This can be converted into finding the pairs with the highest similarities (lowest distance means highest similarity).

Here we can define the similarities between a discovered attribute $\Vec{z}_k$ and a meaningful attribute $\Vec{h}_j$ in terms of their correlations.
Let $\rho(\Vec{z}_k, \Vec{h}_j), \Vec{z}_k \in \mathcal{D}, \Vec{h}_j \in \mathcal{S}$ be the correlation between $\Vec{z}_k$ and $\Vec{h}_j$. 
Then $\rho$ can be defined as:
\begin{equation}
   \rho(\Vec{z}_k, \Vec{h}_j) = \frac{\operatorname{count} (\Vec{z}_k = \Vec{h}_j)}{N},
\end{equation}
\noindent
where $\operatorname{count}$ means the operation which counts how many same elements $\Vec{z}_k$ shares with $\Vec{h}_j$.

Thus, the function $\rho(\Vec{z}_k, \Vec{h}_j)$ can be computed from $\Mat{A}_{\cdot, j}$ and $\Mat{B}_{\cdot,k}$, where 
$\Mat{B}_{\cdot,k}$, $\Mat{A}_{\cdot, j}$ represent the discovered attribute $\Vec{z}_k$ and the meaningful attribute $\Vec{h}_j$ respectively.
Denote $\mathcal{P}$ as the set of M pairs of $\Vec{h}_j \in \mathcal{S}$ and $\Vec{z}_k \in \mathcal{D}$ that have the highest correlation, 
$\mathcal{P} = \{ ( \Vec{h}^1_j, \Vec{z}^1_k ) \cdots ( \Vec{h}^M_j, \Vec{z}^M_k )\}$, 
$\Vec{h}^i_j = \Vec{h}^l_j$ if and only if $i = l$, $\Vec{z}^i_k = \Vec{z}^l_k$.

Therefore the matrix $\Mat{R}^*$ that minimizes~\eqref{eq:joint_prob} is defined, after $\mathcal{P}$ is determined, via:
\begin{equation}
	\label{eq:jp_R}	
	\Mat{R}^*_{j,k} = \left\{ \begin{array}{l}\ 1\ \operatorname{if}\ ( \Vec{h}_j,\Vec{z}_k)  \in  {\cal P}\\
\ 0\ \operatorname{if}\ ( \Vec{h}_j,\Vec{z}_k)  \notin  {\cal P}.
\end{array} \right.
\end{equation}

For the given inputs $\mathcal{D} = \{ \Vec{z}_k \}_{k=1}^K$, ${\mathcal{S}} = \{ \Vec{h}_j \}_{j=1}^{J}$  and $\mathcal{X} = \{ \Vec{x}_i \}_{i=1}^N$, Algorithm~\ref{algo:joint_prob} elaborates the procedures of computing the set $\mathcal{P}$ .
Note that, $( \Vec{h}_j, \cdot )$ and $( \cdot, \Vec{z}_k )$ in step 3 represent all possible pairs containing $\Vec{h}_j$ and $\Vec{z}_k$, respectively.

\begin{algorithm}
  \caption{The proposed greedy algorithm to solve~\eqref{eq:joint_prob}}
  \label{algo:joint_prob}
  \begin{algorithmic}[1]
    \REQUIRE $\mathcal{D} = \{ \Vec{z}_k \}_{k=1}^K$, ${\mathcal{S}} = \{ \Vec{h}_j \}_{s=1}^{J}$  and $\mathcal{X} = \{ \Vec{x}_i \}_{i=1}^N$ 
    \ENSURE $\mathcal{P}$ that contains M pairs that have the highest correlation, where $M = \min ( K, J )$. 
    \STATE $\mathcal{P} \leftarrow \{ \}$    
    \WHILE {$|\mathcal{P}| \le M$}
    	\STATE Find the highest $\rho(\Vec{h}_j, \Vec{z}_k)$ where $( \Vec{h}_j, \cdot ) \notin \mathcal{P}$ and $( \cdot, \Vec{z}_k ) \notin \mathcal{P}$.
    	\STATE $\mathcal{P} = \mathcal{P} \cup (\Vec{h}_j, \Vec{z}_k )$
	\ENDWHILE
  \end{algorithmic}
\end{algorithm}


\section{Attribute Set Meaningfulness Metric}
\label{sec:part3}
Attribute meaningfulness metric is designed to determine which existing automatic attribute discovery method is more likely to discover meaningful attributes.
Moreover, it can provide some insights about how to devise new automatic attribute discovery methods.

In this section we will introduce the Attribute Set Meaningfulness Metric. 
We order our discussion as follows (1) Meaningful subspace interpolation; (2) Selecting meaningful subspace representation and (3) Computing the meaningfulness metric. 

By means of the distance functions $\delta_{\operatorname{jp}}$ and $\delta_{\operatorname{cvx}}$ described in Section~\ref{sec:distance}, we are able to measure how far is the set of discovered attributes $\mathcal{D}$ from the Meaningful Subspace $\mathcal{S}$.
The closer the distance, the more meaningful the set of attributes are.
However, as the relationship between the proposed distances and meaningfulness could be non-linear, the distance may not be easy to interpret.
Furthermore, it is difficult to compare the results between $\delta_{\operatorname{cvx}}$ and $\delta_{\operatorname{jp}}$.

\subsection{Attribute meaningful subspace interpolation}

Our goal is to obtain a metric that is both easy to interpret and able to perform 
comparisons between various distance functions.
Inspired by~\cite{gopalan2014unsupervised}, we apply the subspace interpolation to generate a set of subspaces between Meaningful Subspace and Non-Meaningful Subspace, or Noise Subspace.
Here, we use a set of evenly distributed random attributes to represent the Non-Meaningful Subspace $\mathcal{N}$.

For the purpose of subspace interpolation, the meaningful attribute set $\mathcal{S}$ is divided
into two subsets:

\begin{equation}
\label{eq:set}
 \mathcal{S}^1 \cup \mathcal{S}^2 = \mathcal{S}
\end{equation}

where we consider the set $\mathcal{S}^1$ as the representation of the Meaningful Subspace.
When gradually adding random attributes $\tilde{\mathcal{N}} \in \mathcal{N}$ into $\mathcal{S}^2$, the interpolated sets of subspaces can be obtained.
Here we present the proposition which guarantees that the interpolation is able to generate subspaces between the 
Meaningful Subspace and the Non-Meaningful Subspace.

\begin{proposition}
\label{proposition1}
 Let $\tilde{\mathcal{S}} = \mathcal{S}^2 \cup \tilde{\mathcal{N}}$; when $\tilde{\mathcal{N}} = \{ \}$, the distance $\delta^*$ between 
 $\tilde{\mathcal{S}}$ and $\mathcal{S}^1$ (refers to \eqref{eq:set}) is minimized. However, when 
 $\tilde{\mathcal{N}} -> \mathcal{N}$, the distance between $\tilde{\mathcal{S}}$ and $\mathcal{S}^1$ is asymptotically close to $\delta^*(\mathcal{S}^1,\mathcal{N}; \mathcal{X})$, where $\delta^*$ is the distance function presented previously such as $\delta_{\operatorname{jp}}$ and $\delta_{\operatorname{cvx}}$. More precisely, we can define the relationship as follows:
 \begin{equation}
   \lim_{|\tilde{\mathcal{N}}| \rightarrow \infty} \delta^*(\tilde{\mathcal{S}},\mathcal{S}^1; \mathcal{X}) = \delta^*(\mathcal{N},\mathcal{S}^1; \mathcal{X}).
 \end{equation}
\end{proposition}

\noindent
\textit{Remarks.}
Proposition~\ref{proposition1} basically describes when random attributes are added into $\tilde{\mathcal{S}}$ gradually, the subspace, that is initially close to the Meaningful Subspace $\mathcal{S}^1$, will be more and more distant from $\mathcal{S}^1$. Eventually the subspace will be spanned by random attributes that is asymptotically close to the Non-Meaningful attribute subspace.
While it is easy to prove the above Proposition, we present one version of the proof in the appendix. 

\subsection{Selecting meaningful subspace representation}
\label{sec:subspace_rep}
As discussed in section~\ref{sec:manifold}, enumerating all the meaningful attributes to represent the meaningful attribute subspace is impossible.
We thus use linear combinations of meaningful attributes to approximate the meaningful subspace. 

However, the division of the meaningful attributes into two subsets as suggested in Proposition 4.1 will reduce the subspace spanned to represent the meaningful subspace. 
More specifically, the linear combination of attributes from $\mathcal{S}^1$ may not span the whole meaningful subspace. 
To remedy this, one should carefully select the attributes to form $\mathcal{S}^1$ that can maximize the space spanned by the representation.

Under our proposed approach which is based on the linear reconstruction, the selected meaningful attributes for $\mathcal{S}^1$ should form the bases of the meaningful subspace. 
Here, one way to maximize the space spanned is that to select independent bases.

The attribute independence with respect to the others can be evaluated by how well the attribute can be reconstructed from others.
In addition, one can evaluate the attribute independence from the attribute semantic names. 
For instance, textural attributes such as `metal' may be independent to the other textural attributes such as `grass', `wooden'. 
Therefore, these attributes should be included in the set to represent the meaningful subspace~\ie the set $\mathcal{S}^1$.

In the light of these facts, we propose an approach to perform meaningful subspace representation selection, $\mathcal{S}^1$. 
First, the attribute semantic descriptions are considered. 
Any attributes that are deemed to be independent will be indicated and always put in the set $\mathcal{S}^1$. 
On the second step, we evaluate the attribute independence by applying either $d_{cvx}$ or $d_{jp}$. 
In particular, we use a leave-one-attribute-out scheme which calculates the distance between one attribute to the rest of the attributes. 
We then set the threshold $\alpha$. 
Again, we indicate any attributes having distance more than $\alpha$ and always put them in $\mathcal{S}^1$.
The threshold, $\alpha$ will be one of the parameters which will be determined during the experiments.

\subsection{Computing the meaningfulness metric}
After constructing the meaningful subspace, we can calibrate the attribute set meaningfulness distance by subspace interpolation based on the equivalent distance effect assumption~\cite{bishop2012tensor}.
That is, if the distance of two attribute subspaces to the meaningful subspace are the same, the amount of meaningful contents contained in these two subspaces are the same. 

We denote the distance between $\tilde{\mathcal{S}}$ and the Meaningful Subspace $\mathcal{S}^1$ as $\delta^{\tilde{\mathcal{S}}}$ and the distance between $\mathcal{D}$ and the Meaningful Subspace $\mathcal{S}^1$ as $\delta^{{\mathcal{D}}}$.
After subspace interpolation, we find the subspace $\tilde{\mathcal{S}}$ that makes $\delta^{\tilde{\mathcal{S}}} \approx \delta^{{\mathcal{D}}}$.
Using the equivalent distance effect assumption, if $\delta^{\tilde{\mathcal{S}}} \approx \delta^{{\mathcal{D}}}$, the meaningfulness between $\tilde{\mathcal{S}}$ and $\mathcal{D}$ should be on par with each other.
As $\tilde{\mathcal{S}}$ is defined as a set of meaningful attributes added with additional noise attributes, this representation is able to evaluate the meaningfulness of $\mathcal{D}$.
We can consider this task as an optimization problem as follows:
\begin{equation}
\label{eq:optimization}
 g^* = \mathop{ \arg \min_{|\tilde{\mathcal{N}}|} \left\| {\delta^*(\{\mathcal{S}^2 \cup \tilde{\mathcal{N}\}},{\mathcal{S}^1}; \mathcal{X}) - \delta^*(\mathcal{D},{\mathcal{S}^1}; \mathcal{X})} \right\|_2^2}.
\end{equation}
\noindent
where $g^*$ represents how many minimum number of random attributes required to be added into $\tilde{\mathcal{S}}$ to make $\delta^{\tilde{\mathcal{S}}} \approx \delta^{{\mathcal{D}}}$. 
The above optimization problem can be interpreted as searching for the furthest subspace $\tilde{\mathcal{S}}$ from the Meaningful Subspace in an open sphere with radius $\delta^{{\mathcal{D}}}$.
The above equation can be simply solved by a curve fitting approach. 
In our implementation, we apply the least square approach.

Finally, we denote $\gamma$ as the proposed attribute meaningfulness metric as follows.
\begin{equation}
\label{eq:calibration}
\gamma(\mathcal{D};\mathcal{X}, \mathcal{S})=({1 -{{g^*} \over {|{\mathcal{S}^2}|} + g^*}} ) \times 100.
\end{equation}
\noindent
\textit{Remarks.}
The equation in~\eqref{eq:calibration} determines how many noise/Non-Meaningful attributes are required for a set of automatically discovered attributes to have similar distance as $\delta^{\mathcal{D}}$.
On the other hand, our proposed metric reflects how many meaningful attributes are contained in the attribute set.
A smaller number of Non-Meaningful attributes indicates a more meaningful attribute set overall.

Since different aspects of meaningfulness may be captured by various distance functions, we combine the metric values calculated using both proposed distance functions.
For simplicity, we use a equally weighted summation in this paper: 
$\tilde{\gamma} = \frac{1}{2} \gamma_{\operatorname{cvx}} + \frac{1}{2} \gamma_{\operatorname{jp}}$, as our final metric.

\section{Experiments}
\label{Experiment}
In this part, the efficacy of our approach to measure the meaningfulness of a set of attributes will be first evaluated.
Then the proposed metric is used to evaluate meaningfulness of
the attribute sets generated by various automatic attribute discovery methods such as
PiCoDeS~\cite{bergamo2011picodes} and
Discriminative Binary Codes~(DBC)~\cite{rastegari2012attribute} as well as
some recent hashing methods such as
Iterative Quantization (ITQ)~\cite{gong2011iterative},
Spectral Hashing (SPH)~\cite{weiss2009spectral},
Locality Sensitivity Hashing (LSH)~\cite{Leskovec2009Mining} and
Kernel-Based Supervised Hashing (KSH)~\cite{liu2012supervised}.

The two proposed metrics $\gamma_{\operatorname{jp}}$
~\eqref{eq:joint_prob}, $\gamma_{\operatorname{cvx}}$ ~\eqref{eq:con_hull} and the combined metric $\tilde{\gamma}$ are applied to
compare the meaningfulness of the attributes discovered from the comparative methods
on four attribute datasets: 
(1) a-Pascal a-Yahoo dataset (ApAy)~\cite{Farhadi09describingobjects}; 
(2) Animal with Attributes dataset (AwA)~\cite{Lampert13} and; 
(3) SUN Attribute dataset (ASUN)~\cite{Patterson2012SunAttributes}; 
(4) Unstructured Social Activity Attribute dataset (USAA)~\cite{fu2012attribute}

Finally, our metric will be then compared against a user study and a metric, denoted the MPPCA metric or MPPCA, adapted from semi-supervised attribute discovery method proposed in~\cite{parikh2011interactively}.


\subsection{Datasets and experiment setup}
\noindent
\textbf{a-Pascal a-Yahoo dataset (ApAy)~\cite{Farhadi09describingobjects} --- } comprises two sources: a-Pascal and a-Yahoo.
There are 12,695 cropped images in a-Pascal that are divided into
6,340 for training and 6,355 for testing with 20 categories.
The a-Yahoo set has 12 categories disjoint from the a-Pascal categories.
Moreover, it only has 2,644 test exemplars.
There are 64 attributes provided for each cropped image.
The dataset provides four features for each exemplar: local texture, HOG, edge and color descriptor.
We use the training set for discovering attributes and we perform our study on the test set. More precisely, we consider the test set as the set of images $\mathcal{X}$ defined in \ref{sec:manifold}.

\noindent
\textbf{Animal with Attributes dataset (AwA)~\cite{Lampert13} --- } the dataset contains 35,474 images of 50 animal categories with 85 attribute labels.
There are six features provided in this dataset: HSV color histogram, SIFT~\cite{lowe2004distinctive}, rgSIFT~\cite{van2010evaluating}, PHOG~\cite{bosch2007representing}, SURF~\cite{bay2008speeded} and local self-similarity~\cite{shechtman2007matching}.
The AwA dataset is proposed for studying the zero-shot learning problem.
As such, the training and test categories are disjoint; there are no training images for test categories and vice versa.
More specifically, the dataset contains 40 training categories and 10 test categories.
Similar to the ApAy dataset, we use the training set for discovering attributes and we perform the study in the test set.

\noindent
\textbf{SUN Attribute dataset (ASUN)~\cite{Patterson2012SunAttributes} --- } ASUN is a fine-grained scene classification dataset
consisting of 717 categories (20 images per category) and 14,340 images in total with 102 attributes.
There are four types  of features provided in this dataset: (1) GIST; (2) HOG;
(3) self-similarity and (4) geometric context color histograms~(See \cite{xiao2010sun} for feature and kernel details).
From 717 categories, we randomly select 144 categories for discovering attributes.
As for our evaluation, we randomly select 1,434 images (\ie 10\% of 14,340 images) from the dataset.
It means, in our evaluation, some images may or may not come from the 144 categories used for discovering attributes.

\noindent
\textbf{Unstructured Social Activity Attribute dataset (USAA)~\cite{fu2012attribute} --- } USAA is a relatively novel benchmark  attribute dataset for social activity video classification and annotation.
It is manually annotated with 69 groundtruth attributes from 8 semantic class videos of Columbia Customer Video (CCV) dataset.
There are 100 videos per-class for training and testing respectively.
The annotated attributes can be divided into 5 broad categories: actions,
objects, scenes, sounds, and camera movement.
The 8 classes in the dataset are birthday party, graduation party, music performance, non-music
performance, parade, wedding ceremony, wedding dance and wedding reception.
The SIFT, STIP and MFCC features for all these videos are extracted in the dataset.





For each experiment, we apply the following pre-processing step described in~\cite{bergamo2011picodes}.
We first lift each feature 
into a higher-dimensional space which is three times larger than the original
space.
After the features are lifted, we then apply PCA to reduce the dimensionality of the feature space by 40 percent.
This pre-processing step is crucial for PiCoDeS as it uses lifted feature space to simplify their training scheme while maintaining the information preserved in the Reproducing Kernel Hilbert Space~(RKHS).
Therefore, the method performance will be severely affected when lifted features are not used.



Each method is trained with the training images to discover the attributes.
Then we use the manifold $\mathcal{M}$ w.r.t. the test images for the evaluation.
More precisely, each attribute descriptor is extracted from test images (\ie $\Vec{z}_k, \Vec{z}_k \in \{-1,1\}^N$, where $N$ is the number of test images).
For each dataset, we use the attribute labels from AMT to represent the Meaningful Subspace, $\mathcal{S}$.

We adapted the MPPCA metric from the semi-supervised attribute discovery method proposed in~\cite{parikh2011interactively}. 
In particular, to discover an attribute, the method in~\cite{parikh2011interactively} progressively updates MPPCA model using human feedback. 
In our settings, we directly train MPPCA using attributes found from AMT for each dataset. 
To measure meaningfulness, we compute the posterior probability of the given discovered attribute to the MPPCA model. 
We train MPPCA model using five components and three dimensional subspace for ASUN dataset. 
As for ApAy dataset, we use three components and three dimensional subspace. 
This is because the number of attributes in ApAy dataset is much smaller than ASUN dataset. 
Unless otherwise stated, we follow the experiment settings for MPPCA as described in~\cite{parikh2011interactively}. 
For instance, we employ a threshold on the posterior probability to determine whether an attribute is meaningful. 
The MPPCA metric is the computed by computing the percentage of the attributes deemed as meaningful over the total discovered attributes.

\subsection{Do $\delta_{\operatorname{cvx}}$ and $\delta_{\operatorname{jp}}$ measure meaningfulness?}
In this experiment, we evaluate whether the proposed approach really does measure the meaningfulness on a set of automatically discovered attributes. 
One of the key assumptions in our proposal is that the distance between the Meaningful Subspace and the given attribute set $\mathcal{D}$ reflects the meaningfulness of a set of attribute. 
More specifically, if the distance is small, it is assumed that the attribute set is potentially meaningful and vice versa.
Aiming to evaluate that, we construct two sets of attributes, respectively with meaningful and non-meaningful attributes and observe their distances to the meaningful subspace.

As to the meaningful attribute set, we follow the methods used in Section~\ref{sec:part3}.
Providing manually labelled attribute set $\mathcal{S}$, here denoted the AMT attribute set~\footnote{As mentioned before, attributes discovered from the AMT procedure are considered meaningful.
For the sake of clarity, we call these attributes AMT attributes.}, in each dataset,
we follow the approach used in Section~\ref{sec:part3} to divide the set into two subsets $\mathcal{S}^1
\cup \mathcal{S}^2 = \mathcal{S}$ where
$\mathcal{S}^1$ represents the
Meaningful Subspace and $\mathcal{S}^2$ is considered as a set of
discovered attributes (\ie $\mathcal{D} = \mathcal{S}^2$).
Unequivocally, the attributes in $\mathcal{S}^2$ should be meaningful as they are manually labelled by human annotator. Thus, we name $\mathcal{S}^2$ as the $\textit{MeaningfulAttributeSet}$.

For non-meaningful attribute set, we create this set by
randomly generating the attributes. 
As described in Section~\ref{sec:part3}, we generate a finite set of random attributes denoted by
$\tilde{\mathcal{N}}$.
We name this set as \textit{NonMeaningfulAttributeSet} since it is non-meaningful and should have significantly larger distance to the Meaningful Subspace.

\begin{figure}[htbp]
	\centering
		\subfigure[]{
		\label{fig:dataset:a} 
		\includegraphics[width=1\textwidth]{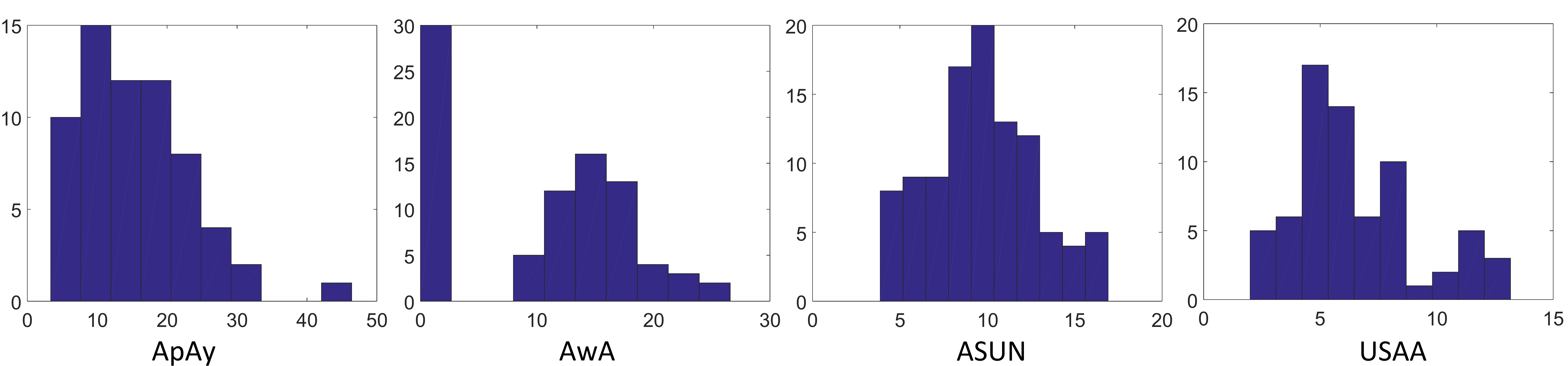}}
		
		\subfigure[]{
		\label{fig:dataset:b} 
		\includegraphics[width=1\textwidth]{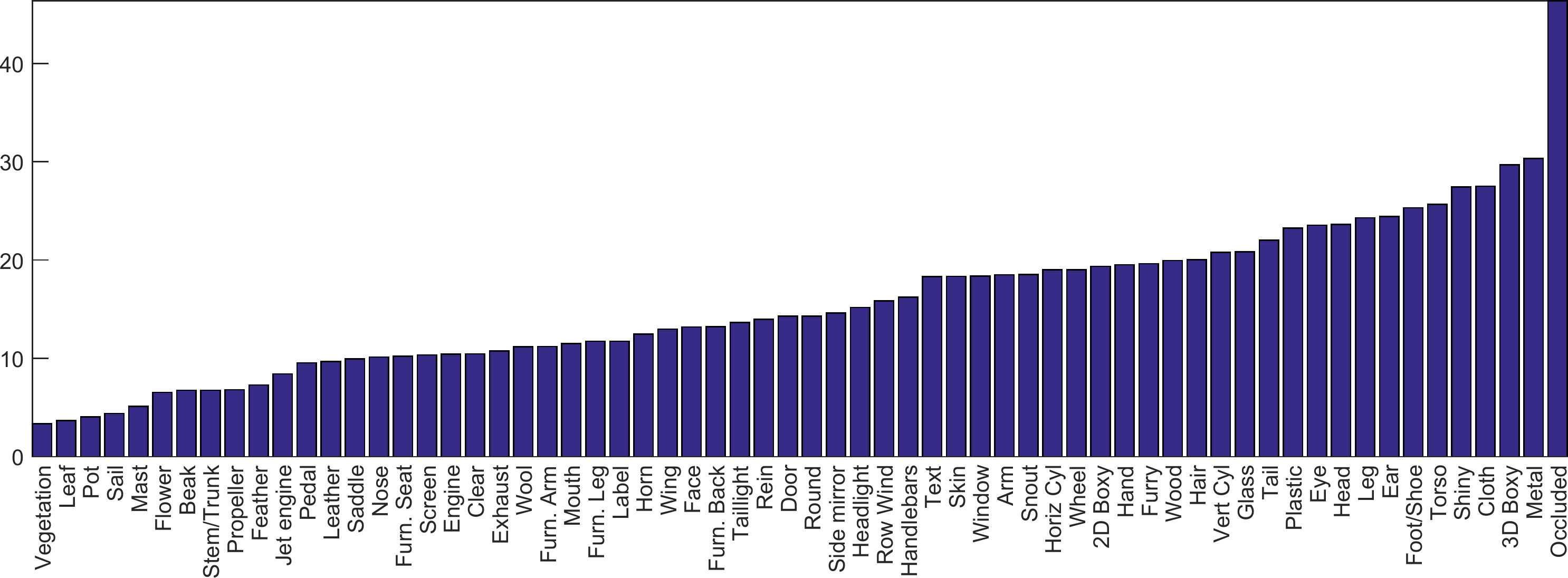}}
		\caption{The reconstruction error analysis on datasets. In (a), the horizontal axis represents the reconstruction error value, the vertical axis represents the frequency of the attributes which fall into the range of reconstruction error. In (b), the horizontal axis depicts the name of each attribute, the vertical axis represents the reconstruction error value.}
\end{figure}

Recall that the Meaningful Subspace $\mathcal{S}^1$ needs to be carefully selected to maximize the meaningful subspace spanned. 
However, to show the efficacy of our proposed selection, we first randomly select $\mathcal{S}^1$. 
Then, on the second experiment, we apply our proposed selection approach.
To perform our proposed selection approach, we must evaluate the independence of each AMT attribute via analysing its attribute name and computing its individual reconstruction error. 
We will always put independent attributes in $\mathcal{S}^1$. 
In other words, let $\mathcal{\hat{S}}^1$ be the set of AMT attributes marked as independent attributes. 
Then, the set $\mathcal{S}$ is divided into $\mathcal{S}^1$ and $\mathcal{S}^2$ such that, $\mathcal{\hat{S}}^1$ will be always in $\mathcal{S}^1$.
In this case, we still randomly divide $\mathcal{S}$ with a constraint that the $\mathcal{\hat{S}}^1$ should always be in the set $\mathcal{S}^1$.
As previously described, a leave-one-attribute-out scheme is used to determine the independence of an AMT attribute with respect to the rest of AMT attribute set. 
Fig.~\ref{fig:dataset:a} presents the result of this analysis.

As we can see, the reconstruction error of ApAy dataset are, in general, much larger than the other datasets.
We conjecture that this might caused by the fact that the other three  datasets are all fine-grain classification datasets, however ApAy is proposed for addressing the general classification problem. 
This means the attributes provided in this dataset are more likely to be independent as they are used to describe a wide variety of classes.
For further inspection, we take the ApAy dataset and present the results in the perspective of semantic reasoning of the attribute names. 
In Fig.~\ref{fig:dataset:b}, we plot the reconstruction error for each attribute in ascending order. 
As we can see, the attributes with low reconstruction errors are often more likely to be correlated (\ie less independent).
For example, images such as the ``leaf'' and ``pot'' can reconstruct the ``vegetation'' attribute, ``Sail'' and ``Mast'' are often present together in the sailing scenes. 
However, the attributes with high reconstruction error are more likely to be independent (i.e. less correlated) such as shape-related attribute ``3D Boxy'', material-related attribute ``Metal'' and especially ``Occluded''.
It is worthy to note that we only show the analysis using reconstruction error with convex hull regularization.
The same findings are also exhibited when the $\ell_0$ regularization is used.

To reasonably determine the parameter $\alpha$ (See~\ref{sec:subspace_rep}), we average the highest reconstruction error scores from the other three fine-grain datasets.
This gives us a value $\alpha = 18.89$.
That means we consider any attributes in the datasets with error above $\alpha$ to be independent attributes.
Thus the 22 independent attributes with highest reconstruction error can be put into the meaningful attribute subset $\mathcal{S}^1$ for better approximation of meaningful attribute subspace.
The rest of the attributes are still pooled and randomly selected.
Table~\ref{tab:apay_threshold} shows the results with and without the proposed selection strategy.
As we can see, after applying the selection, the MeaningfulAttributeSet which is always considered as meaningful, exhibits the lowest reconstruction error.
Other methods almost remain the same with little random perturbation that indicates for the automatic attribute discovery methods our metric is quite stable.
Again, we note that we use $\delta_{\operatorname{cvx}}$ and similar results are also found when using $\delta_{\operatorname{jp}}$.

\begin{table}[htbp]
  \centering
  \caption{Comparisons of reconstruction error results on ApAy dataset with and without the selection strategy.}
    \begin{tabular}{ccc}
    \toprule
          & With selection & Without selection \\
    \midrule
    PiCoDeS & 12.65 & 12.52 \\
    DBC   & 48.97 & 49.66 \\
    ITQ   & 50.73 & 51.70 \\
    SPH   & 48.91 & 49.79 \\
    LSH   & 52.17 & 53.14 \\
    KSH   & 38.34 & 38.66 \\
    NonMeaningful & 53.20 & 54.34 \\
    Meaningful & 12.09 & 18.29 \\
    \bottomrule
    \end{tabular}%
  \label{tab:apay_threshold}%
\end{table}%

Now, we are ready to discuss the evaluation of our proposed approach to determine whether our approach can measure the attribute meaningfulness.
In order to do that, we first perform the subspace interpolation of all the attribute set discovered by the methods.
To perform the subspace interpolation, the random attributes are progressively added to the set of attributes from each method.
By doing this, we can evaluate if the distance to Meaningful Subspace is
enlarged when we progressively increase the number of non-meaningful attributes.

\begin{figure*}
    \centering
    \includegraphics[width=1\linewidth]{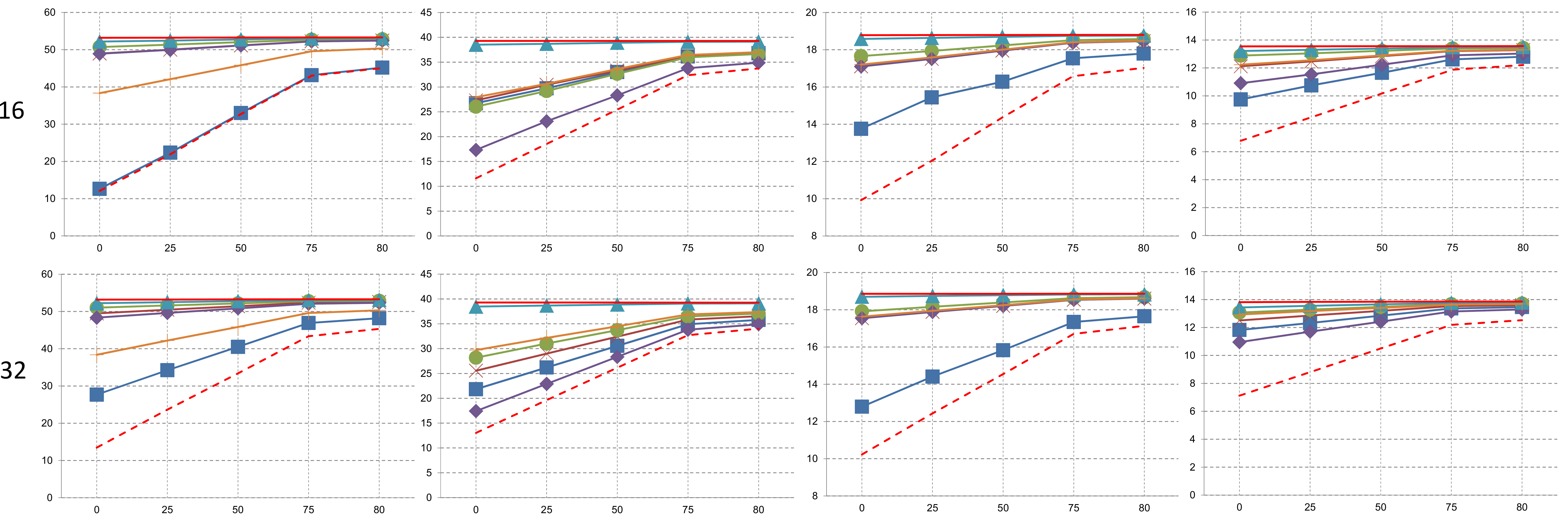}  \\
    \includegraphics[width=1\linewidth]{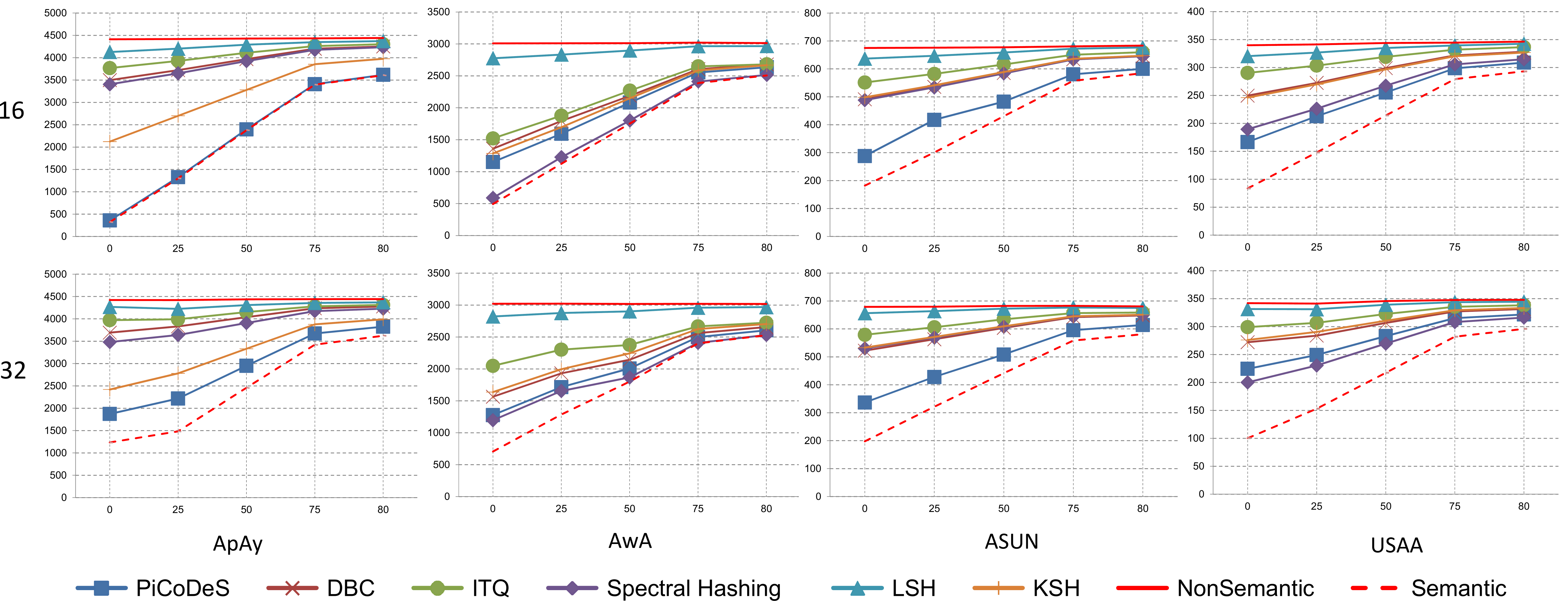}
    
       \caption{Validation of attribute meaningfulness measurement by reconstruction error $\delta_{\operatorname{cvx}}$ (first and second rows) and $\delta_{\operatorname{jp}}$ (third and fourth rows).
       In each subfigure, the horizontal axis represents the percentage of noise attributes, the vertical axis means the reconstruction error values.
        As we can see, both distances become larger when more random/non-meaningful attributes are added.
        MeaningfulAttributeSet has the closest distance to the Meaningful Subspace and NonMeaningfulAttributeSet always has the largest distance. 
        Here, each method is configured to discover 16 and 32 attributes.
		The smaller the $\delta$, the more meaningful the attribute set is.
        }
             \label{fig:noise}
\end{figure*}

Fig.~\ref{fig:noise} presents the evaluation results. 

Here we present the reconstruction error results for all 4 datasets where 16 and 32 attributes are discovered by the methods respectively.
Recall that although we carefully selected $\mathcal{S}^1$, the set $\mathcal{S}$ is still randomly divided. 
We produced the results shown in Fig.~\ref{fig:noise} by repeating the random division of $\mathcal{S}^1$ 100 times and calculated the average distance.
The detail results are also shown in Table~\ref{tab:std}.
We note that no matter how $\mathcal{S}^1$ is selected, our method is relatively stable.

\begin{table}[htbp]
  \centering
  \caption{
  The table of values in reconstruction errors with standard deviation analysis.
  Letter E conventionally represents 'times ten raised to the power of'
  }
  \scalebox{0.5}{
    \begin{tabular}{c|c|r@{}lr@{}lr@{}lr@{}l|r@{}lr@{}lr@{}lr@{}l}
    \toprule
          &       & \multicolumn{8}{c}{cvx}                                       & \multicolumn{7}{c}{jp}                                &  \\
    \midrule
          &       & \multicolumn{2}{c}{ApAy} & \multicolumn{2}{c}{AwA} & \multicolumn{2}{c}{ASUN} & \multicolumn{2}{c}{USAA} & \multicolumn{2}{c}{ApAy} & \multicolumn{2}{c}{AwA} & \multicolumn{2}{c}{ASUN} & \multicolumn{2}{c}{USAA} \\
    \multirow{8}[1]{*}{16} & \multicolumn{1}{c|}{PiCoDeS} & 12.65 & \multicolumn{1}{l}{$\pm$9.57E-02} & 26.73 & \multicolumn{1}{l}{$\pm$7.10E-02} & 13.76 & \multicolumn{1}{l}{$\pm$2.33E-02} & 9.74  & \multicolumn{1}{l|}{$\pm$5.57E-02} & 358.61 & \multicolumn{1}{l}{$\pm$1.53E+01} & 1150.76 & \multicolumn{1}{l}{$\pm$1.71E+01} & 287.95 & \multicolumn{1}{l}{$\pm$8.80E-01} & 166.57 & \multicolumn{1}{l}{$\pm$3.17E+00} \\
          & \multicolumn{1}{c|}{DBC} & 48.97 & \multicolumn{1}{l}{$\pm$1.78E-04} & 27.30 & \multicolumn{1}{l}{$\pm$2.04E-01} & 17.10 & \multicolumn{1}{l}{$\pm$2.19E-02} & 12.12 & \multicolumn{1}{l|}{$\pm$1.14E-01} & 3499.92 & \multicolumn{1}{l}{$\pm$2.93E+00} & 1358.82 & \multicolumn{1}{l}{$\pm$2.32E+01} & 491.13 & \multicolumn{1}{l}{$\pm$1.01E+00} & 249.69 & \multicolumn{1}{l}{$\pm$3.39E+00} \\
          & \multicolumn{1}{c|}{ITQ} & 50.73 & \multicolumn{1}{l}{$\pm$3.15E-04} & 26.04 & \multicolumn{1}{l}{$\pm$3.51E-01} & 17.66 & \multicolumn{1}{l}{$\pm$4.85E-02} & 12.88 & \multicolumn{1}{l|}{$\pm$6.79E-02} & 3768.67 & \multicolumn{1}{l}{$\pm$4.66E+00} & 1519.80 & \multicolumn{1}{l}{$\pm$6.30E+01} & 551.51 & \multicolumn{1}{l}{$\pm$3.13E+00} & 290.29 & \multicolumn{1}{l}{$\pm$1.03E+00} \\
          & \multicolumn{1}{c|}{SPH} & 48.91 & \multicolumn{1}{l}{$\pm$1.26E-03} & 17.32 & \multicolumn{1}{l}{$\pm$1.12E-01} & 17.10 & \multicolumn{1}{l}{$\pm$1.85E-02} & 10.91 & \multicolumn{1}{l|}{$\pm$3.40E-02} & 3406.41 & \multicolumn{1}{l}{$\pm$3.22E+00} & 588.92 & \multicolumn{1}{l}{$\pm$3.38E+01} & 488.34 & \multicolumn{1}{l}{$\pm$9.30E-01} & 189.45 & \multicolumn{1}{l}{$\pm$8.48E-01} \\
          & \multicolumn{1}{c|}{LSH} & 52.17 & \multicolumn{1}{l}{$\pm$1.15E-09} & 38.51 & \multicolumn{1}{l}{$\pm$7.64E-03} & 18.58 & \multicolumn{1}{l}{$\pm$3.91E-02} & 13.22 & \multicolumn{1}{l|}{$\pm$9.34E-02} & 4127.82 & \multicolumn{1}{l}{$\pm$6.09E+00} & 2775.42 & \multicolumn{1}{l}{$\pm$3.59E+00} & 636.73 & \multicolumn{1}{l}{$\pm$1.35E+00} & 320.39 & \multicolumn{1}{l}{$\pm$1.53E+00} \\
          & \multicolumn{1}{c|}{KSH} & 38.34 & \multicolumn{1}{l}{$\pm$1.53E-02} & 27.94 & \multicolumn{1}{l}{$\pm$9.73E-02} & 17.21 & \multicolumn{1}{l}{$\pm$1.91E-02} & 12.23 & \multicolumn{1}{l|}{$\pm$6.70E-02} & 2122.98 & \multicolumn{1}{l}{$\pm$4.50E+00} & 1285.05 & \multicolumn{1}{l}{$\pm$9.89E+00} & 498.27 & \multicolumn{1}{l}{$\pm$1.48E+00} & 246.14 & \multicolumn{1}{l}{$\pm$1.40E+00} \\
          & \multicolumn{1}{c|}{NonMeaningful} & 53.20 & \multicolumn{1}{l}{$\pm$1.52E-09} & 39.27 & \multicolumn{1}{l}{$\pm$4.32E-14} & 18.78 & \multicolumn{1}{l}{$\pm$2.75E-02} & 13.55 & \multicolumn{1}{l|}{$\pm$1.21E-01} & 4411.74 & \multicolumn{1}{l}{$\pm$7.45E-01} & 3008.71 & \multicolumn{1}{l}{$\pm$8.16E-01} & 675.03 & \multicolumn{1}{l}{$\pm$3.84E-01} & 339.79 & \multicolumn{1}{l}{$\pm$2.52E+00} \\
          & \multicolumn{1}{c|}{Meaningful} & \textbf{12.09} & \multicolumn{1}{l}{\textbf{$\pm$1.09E+00}} & \textbf{11.60} & \multicolumn{1}{l}{\textbf{$\pm$2.08E+00}} & \textbf{9.92}  & \multicolumn{1}{l}{\textbf{$\pm$6.22E-01}} & \textbf{6.78}  & \multicolumn{1}{l|}{\textbf{$\pm$6.27E-01}} & \textbf{322.44} & \multicolumn{1}{l}{\textbf{$\pm$9.29E+01}} & \textbf{492.54} & \multicolumn{1}{l}{\textbf{$\pm$1.06E+02}} & \textbf{182.04} & \multicolumn{1}{l}{\textbf{$\pm$2.20E+01}} & \textbf{83.54} & \multicolumn{1}{l}{\textbf{$\pm$1.44E+01}} \\
          \hline
    \multirow{8}[1]{*}{32} & \multicolumn{1}{c|}{PiCoDeS} & 27.67 & \multicolumn{1}{l}{$\pm$1.12E-01} & 21.84 & \multicolumn{1}{l}{$\pm$5.42E-02} & 12.79 & \multicolumn{1}{l}{$\pm$4.39E-02} & 11.82 & \multicolumn{1}{l|}{$\pm$1.49E-01} & 1872.33 & \multicolumn{1}{l}{$\pm$3.95E+01} & 1277.40 & \multicolumn{1}{l}{$\pm$6.75E+01} & 336.71 & \multicolumn{1}{l}{$\pm$4.75E+00} & 224.44 & \multicolumn{1}{l}{$\pm$1.96E+00} \\
          & \multicolumn{1}{c|}{DBC} & 49.50 & \multicolumn{1}{l}{$\pm$1.25E-04} & 25.56 & \multicolumn{1}{l}{$\pm$2.69E-01} & 17.55 & \multicolumn{1}{l}{$\pm$5.88E-02} & 12.50 & \multicolumn{1}{l|}{$\pm$3.02E-01} & 3691.58 & \multicolumn{1}{l}{$\pm$7.46E+00} & 1563.89 & \multicolumn{1}{l}{$\pm$6.90E+01} & 522.29 & \multicolumn{1}{l}{$\pm$1.61E+00} & 272.04 & \multicolumn{1}{l}{$\pm$2.89E+00} \\
          & \multicolumn{1}{c|}{ITQ} & 51.02 & \multicolumn{1}{l}{$\pm$2.57E-04} & 28.18 & \multicolumn{1}{l}{$\pm$4.31E-01} & 17.92 & \multicolumn{1}{l}{$\pm$1.09E-01} & 13.07 & \multicolumn{1}{l|}{$\pm$2.09E-01} & 3971.37 & \multicolumn{1}{l}{$\pm$1.22E+01} & 2048.51 & \multicolumn{1}{l}{$\pm$6.32E+01} & 578.97 & \multicolumn{1}{l}{$\pm$3.25E+00} & 298.99 & \multicolumn{1}{l}{$\pm$1.66E+00} \\
          & \multicolumn{1}{c|}{SPH} & 48.35 & \multicolumn{1}{l}{$\pm$1.62E-03} & 17.43 & \multicolumn{1}{l}{$\pm$1.23E-01} & 17.54 & \multicolumn{1}{l}{$\pm$4.82E-02} & 10.95 & \multicolumn{1}{l|}{$\pm$1.08E-01} & 3480.28 & \multicolumn{1}{l}{$\pm$2.39E+01} & 1196.88 & \multicolumn{1}{l}{$\pm$1.01E+02} & 530.47 & \multicolumn{1}{l}{$\pm$1.45E+00} & 200.20 & \multicolumn{1}{l}{$\pm$2.80E+00} \\
          & \multicolumn{1}{c|}{LSH} & 52.22 & \multicolumn{1}{l}{$\pm$3.37E-09} & 38.48 & \multicolumn{1}{l}{$\pm$1.12E-02} & 18.70 & \multicolumn{1}{l}{$\pm$9.51E-02} & 13.43 & \multicolumn{1}{l|}{$\pm$2.99E-01} & 4268.15 & \multicolumn{1}{l}{$\pm$9.82E+00} & 2822.03 & \multicolumn{1}{l}{$\pm$1.02E+01} & 656.52 & \multicolumn{1}{l}{$\pm$2.36E+00} & 331.45 & \multicolumn{1}{l}{$\pm$1.54E+00} \\
          & \multicolumn{1}{c|}{KSH} & 38.37 & \multicolumn{1}{l}{$\pm$3.52E-02} & 29.71 & \multicolumn{1}{l}{$\pm$1.29E-01} & 17.63 & \multicolumn{1}{l}{$\pm$4.34E-02} & 12.93 & \multicolumn{1}{l|}{$\pm$2.68E-01} & 2419.02 & \multicolumn{1}{l}{$\pm$1.52E+01} & 1637.47 & \multicolumn{1}{l}{$\pm$5.03E+01} & 533.15 & \multicolumn{1}{l}{$\pm$1.46E+00} & 276.13 & \multicolumn{1}{l}{$\pm$1.95E+00} \\
          & \multicolumn{1}{c|}{NonMeaningful} & 53.19 & \multicolumn{1}{l}{$\pm$6.56E-10} & 39.28 & \multicolumn{1}{l}{$\pm$1.18E-04} & 18.86 & \multicolumn{1}{l}{$\pm$8.02E-02} & 13.82 & \multicolumn{1}{l|}{$\pm$3.73E-01} & 4421.49 & \multicolumn{1}{l}{$\pm$1.13E+00} & 3020.82 & \multicolumn{1}{l}{$\pm$1.69E+00} & 678.89 & \multicolumn{1}{l}{$\pm$6.81E-01} & 342.10 & \multicolumn{1}{l}{$\pm$1.47E+00} \\
          & \multicolumn{1}{c|}{Meaningful} & \textbf{13.45} & \multicolumn{1}{l}{\textbf{$\pm$5.79E-01}} & \textbf{13.04} & \multicolumn{1}{l}{\textbf{$\pm$1.51E+00}} & \textbf{10.23} & \multicolumn{1}{l}{\textbf{$\pm$5.13E-01}} & \textbf{7.12}  & \multicolumn{1}{l|}{\textbf{$\pm$4.13E-01}} & \textbf{1234.20} & \multicolumn{1}{l}{\textbf{$\pm$9.14E+01}} & \textbf{706.19} & \multicolumn{1}{l}{\textbf{$\pm$9.04E+01}} & \textbf{197.50} & \multicolumn{1}{l}{\textbf{$\pm$1.50E+01}} & \textbf{100.29} & \multicolumn{1}{l}{\textbf{$\pm$7.55E+00}} \\
    \bottomrule
    \end{tabular}%
  \label{tab:std}%
  }
\end{table}%
As we can see from the results that the \textit{MeaningfulAttributeSet} has the closest
distance to the Meaningful Subspace for both
distances $\delta_{\operatorname{cvx}}$ and
$\delta_{\operatorname{jp}}$ on all datasets. 
As expected, the \textit{NonMeaningfulAttributeSet} has the largest distance compared with the others. 
In addition, when the random attributes are progressively added, the distance between the Meaningful Subspace and the sets of attributes discovered by each method increases.
These results indicate that the proposed approach could be used to measure the meaningfulness of a set of attributes. 
Moreover, they also give a strong
indication that there is a \textit{shared structure} between meaningful attributes.

\subsection{Attribute co-occurrence matrix analysis}
\begin{figure*}
    \centering
    \includegraphics[width=1\linewidth]{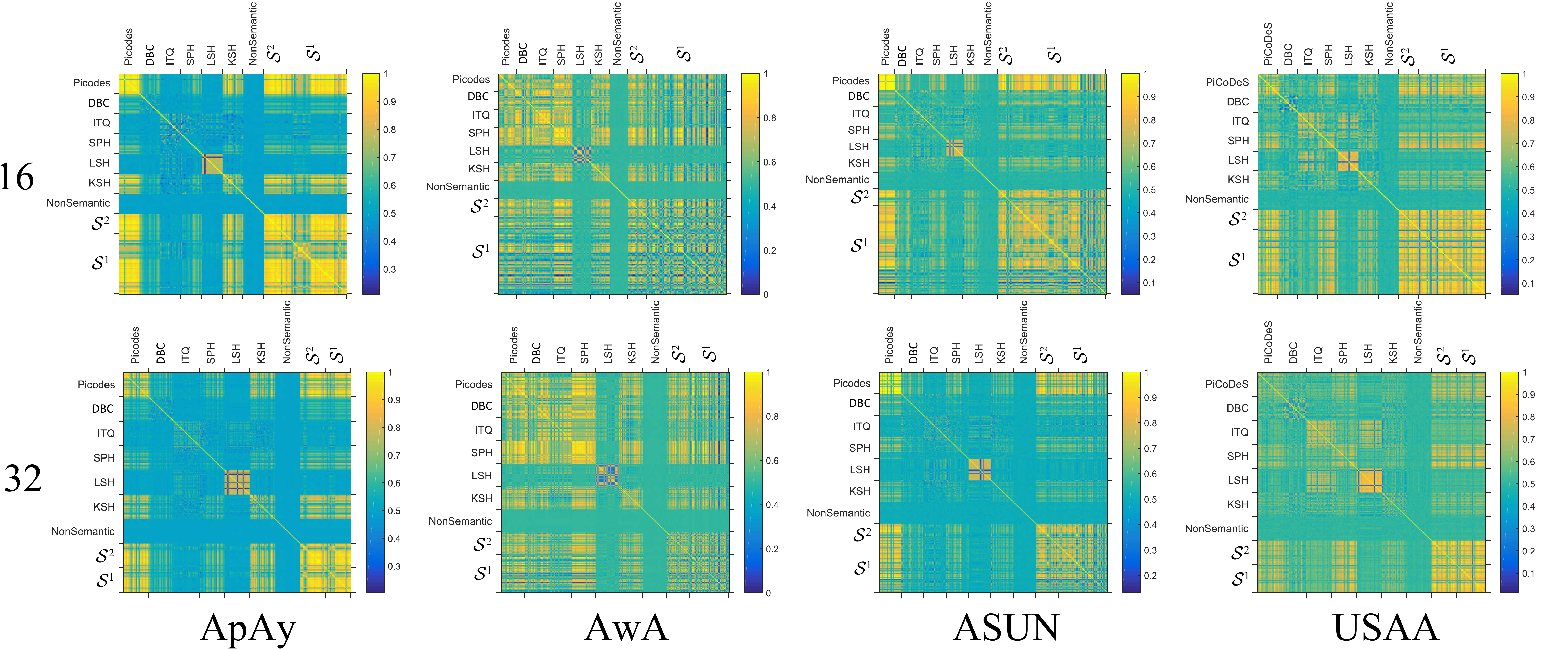} 
    
       \caption{Visualization of co-occurrence matrix, the color bar represents the value of joint probability.        
       The range between two ticks represents the attributes from each method.
       The first row shows the result for each dataset when each method is configured to discover 16 attributes.
       The second row shows the result for each dataset when each method is configured to discover 32 attributes.
        }
             \label{fig:matrix}
\end{figure*}

For further inspection, we also perform the co-occurrence matrix analysis on the attributes discovered by each method and the AMT attributes~\ie $\mathcal{S}^1$ and $\mathcal{S}^2$.
The results are shown in Fig.~\ref{fig:matrix}.

The co-occurrence matrix figures represent the visualization of joint probability between the discovered attributes from each method and the AMT attributes, which are considered as meaningful. 
As we can see in the figure, almost in every dataset, the highest joint probability is achieved between $\mathcal{S}^1$ and $\mathcal{S}^2$.
The trend is quite obvious in ApAy dataset and ASUN dataset and USAA dataset.
However, the trend does not look apparent in AwA dataset. 
We conjecture this could be that there are many attributes in $\mathcal{S}$ that are independent. 
The attributes in AwA dataset are class-level \ie each sample in same class has the same attribute representation.
Therefore, in order to guarantee the discriminative power between classes, the attributes may be chosen to reflect different aspects of classes, thus they could have lower joint probability.
We note that the supervised attribute learning methods such as PiCoDeS, DBC and KSH also have comparable high probability with the AMT attributes.
Another finding is that the attribute generated by the LSH method tend to have high joint probability with each other.
This may be due to the simple linear projection of the data feature matrix in generating the final attribute representation.
Generally, the results are consistent with the previous experiment, which further indicate the capability of our approach to capture the attribute meaningfulness.

\subsection{Attribute set meaningfulness evaluation using $\delta_{\operatorname{cvx}}$ and $\delta_{\operatorname{jp}}$}
\label{sec:att-eval}
In this section, the meaningfulness is evaluated by $\delta_{\operatorname{cvx}}$ and $\delta_{\operatorname{jp}}$ for the set of attributes automatically discovered by various comparative methods in the literature. 
For that purpose, all manually labelled attributes from AMT in each dataset are used as the representation of the Meaningful Subspace.
Then each method is configured to discover 16, 32, 64 and 128 attributes.

\begin{figure*}
    \centering
    \includegraphics[width=1\linewidth]{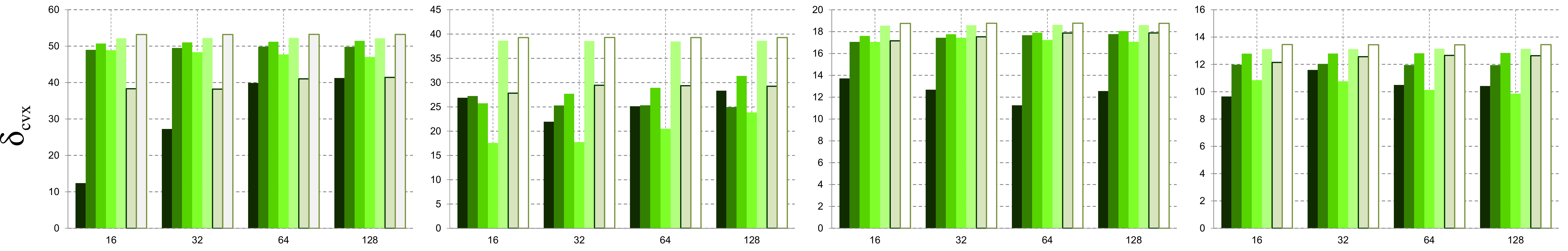}  \\
    \includegraphics[width=1\linewidth]{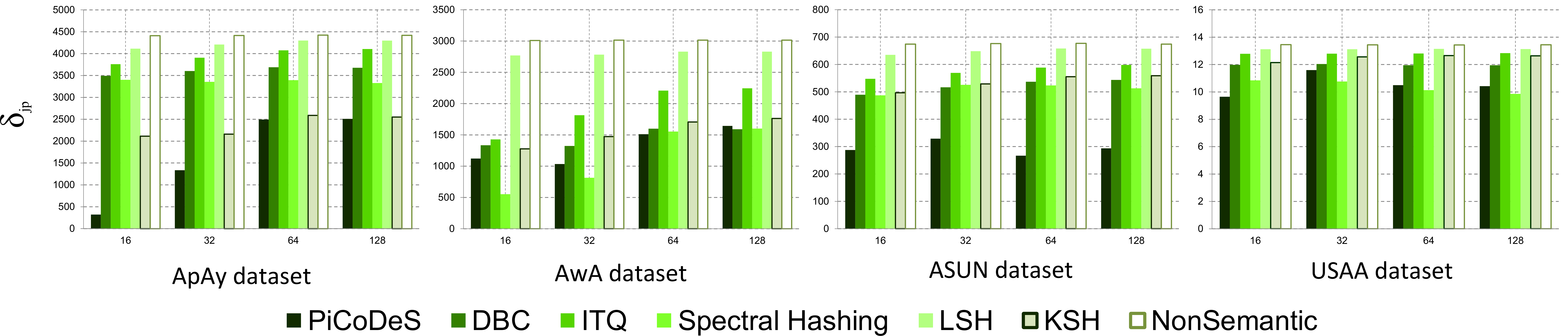}
    \caption{\textit{Attribute meaningfulness} comparisons between different methods on variant number of discovered attributes. The first row reports the results using $\delta_{\operatorname{cvx}}$ and the second row reports the results using $\delta_{\operatorname{jp}}$.
The smaller the $\delta$, the more meaningfulness.}
             \label{fig:recon}
\end{figure*}

Fig.~\ref{fig:recon} reports the evaluation results on all
datasets. 
It is noteworthy to mention that both the proposed distances $\delta_{\operatorname{cvx}}$  and $\delta_{\operatorname{jp}}$  are not calibrated and scaled; making it difficult to perform in-depth evaluation. However, we still can evaluate the results in terms of the method rank ordering (\ie which method  takes first place and which comes the second).

PiCoDeS has the smallest distance in various number of attributes extracted on most of the datasets. 
PiCoDeS uses category labels and employs a max-margin framework to jointly learn the category classifier and attribute descriptor in an attempt to maximize the discriminative power of the descriptor. 
In other words, the goal of PiCoDeS is to discover a set of attributes which can discriminate between categories.

DBC is also developed under the max-margin framework to extract meaningful attributes as PiCoDeS. 
However, compared with PiCoDeS, DBC discovers less meaningful attributes.
We conjecture the reason could be DBC learns the whole attribute descriptors for each category simultaneously unlike PiCoDeS that learns the attribute individually.
This scheme will inevitably emphasize category discrimination of attributes rather than preserving the meaningfulness of individual attribute.
Note that here we do not suggest that DBC is not able to discover meaningful attributes,
rather, PiCoDeS may find more meaningful attributes.
Therefore, our finding does not contradict the results presented in the DBC original paper \cite{rastegari2012attribute}.

Another observation from the results of SPH indicate that it is able to discover meaningful attributes.
SPH is aimed to discover binary codes via a graph embedding approach preserving the local neighborhood structure.
One possible explanation could be that when two images belong to the same class,
they should share more attributes indicating a shorter distance
between them in the binary space, and vice versa.

Although ITQ aims to learn similarity preserving binary descriptor,
it has a larger distance than SPH, DBC and PiCoDeS.
The reason may be the way ITQ learns the binary descriptor which mainly relies on the global information of the data distribution.
In other word, the algorithm minimizes the quantization error of the mapping data to the vertices of a zero centered binary hypercube suggesting that only global information by itself might not be sufficient to discover meaningful attributes.


As expected, the attribute sets from LSH have the largest distances to the Meaningful Subspace (\ie least meaningfulness).
LSH uses random hyperplanes to project a data point into the binary
space.
Therefore, the consistent identifiable visual concepts are hardly 
presented in the positive images.

In summary, two recipes could be derived from the current results that could be
significant for the future automatic attribute discovery method design: the method should attempt to preserve local neighborhood structure as well as to consider the discriminative power of attributes.

\subsection{Attribute set meaningfulness calibration using the proposed meaningfulness metric}
As described in section~\ref{sec:att-eval}, the distance between
attribute sets and the meaningful subspace have some limitations preventing us to perform in-depth analysis.
Quantitative comparisons between different methods are more desirable in analysis of attribute meaningfulness.
Here we report the meaningfulness metric results.

\noindent
\begin{figure}[htb]
    \centering
    \includegraphics[width=0.475\linewidth]{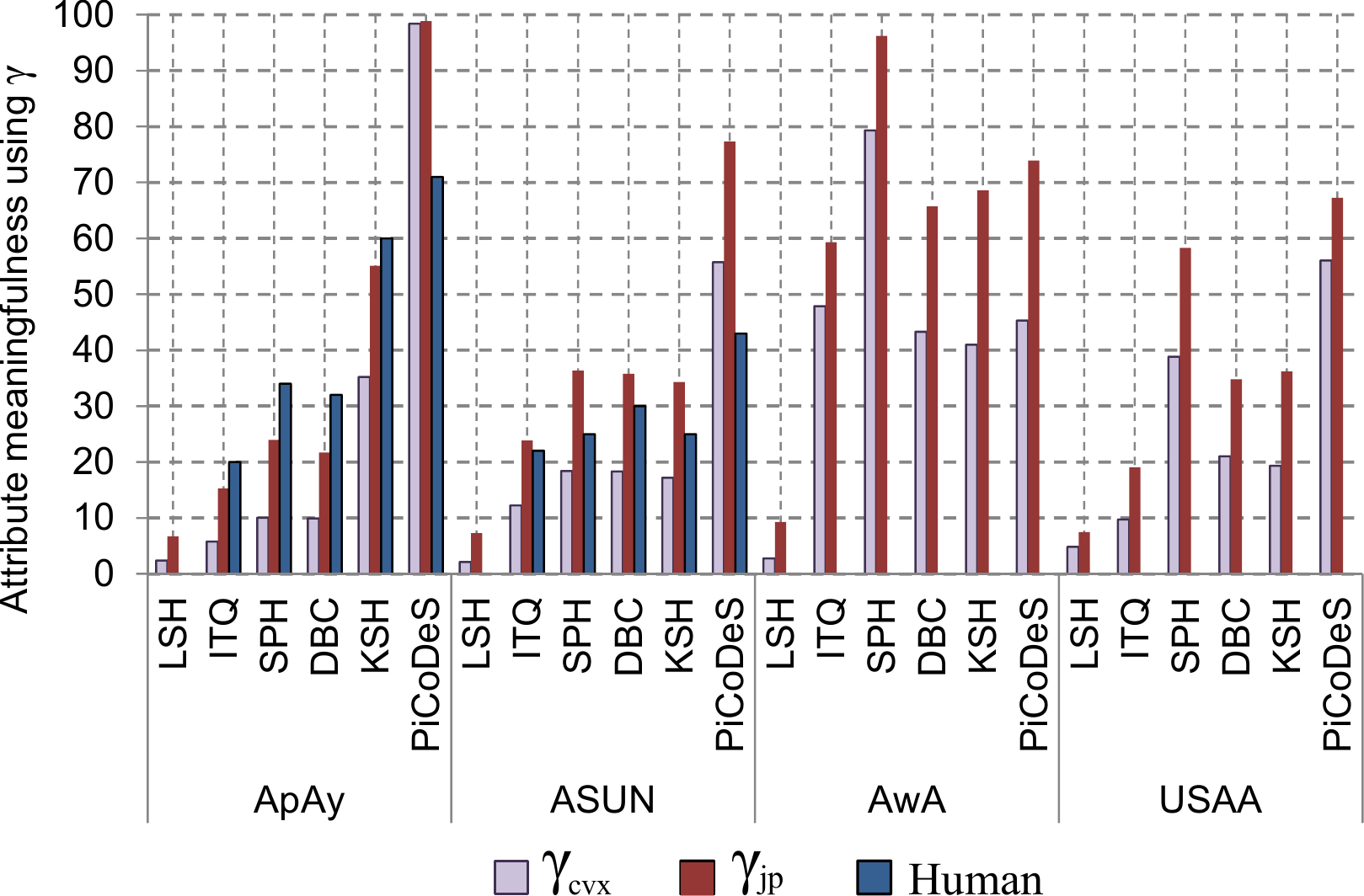}
    \includegraphics[width=0.475\linewidth]{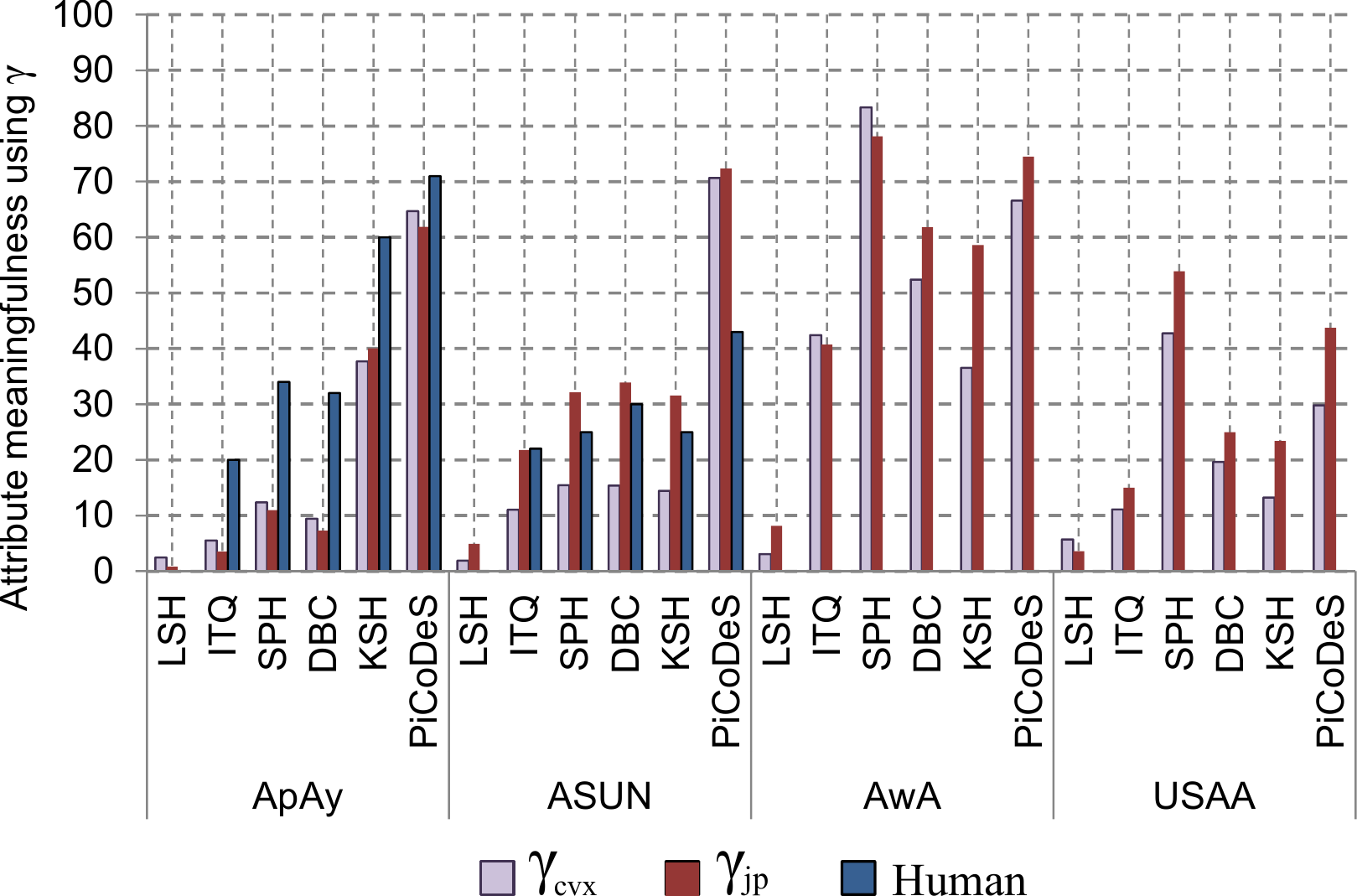}
        \caption{Comparisons of various methods using the proposed meaningfulness metric as well as human study results. 
        Each method is set to discover 16 and 32 attributes. 
        The higher the more meaningful. 
        Human study is not conducted for AWA dataset as special zoology knowledge is required, nor for USAA dataset for inconvenience to display and subjectiveness bias problem.
	    The human results for LSH method are 0 for ApAy and ASUN datasets.
}
             \label{fig:calibration}
\end{figure}

As shown in \eqref{eq:calibration}, we apply $\gamma_{\operatorname{cvx}}$ and $\gamma_{\operatorname{jp}}$ on the datasets and for each method by calibrating the proposed distances $\delta_{\operatorname{cvx}}$ and $\delta_{\operatorname{jp}}$.

The results are shown in Fig.~\ref{fig:calibration} when each method is configured to discover 16 and 32 attributes.
The rank orders of the methods are the almost the same with similar values in most tests by metric $\gamma_{\operatorname{cvx}}$ and $\gamma_{\operatorname{jp}}$, with two exceptions in ASUN dataset. 
This can be explained to the fact that each metric captures a
different aspect of attribute meaningfulness. 
The proposed
$\gamma_{\operatorname{cvx}}$ captures a one-to-many relationship while $\gamma_{\operatorname{jp}}$
evaluates the one-to-one relationship. 
Then the equal weighted metric score $\tilde{\gamma}$ is applied for further analysis.

A user study is also conducted on the attributes discovered by each method.
Since AwA requires experts in animal studies and USAA is a quite large video dataset whose complex social group activities are likely to cause subjectivity bias, we only use ApAy and ASUN dataset for the user study. 

The study collected over 100 responses for each number of discovered attributes. 
In each response, there are positive and negative images presented from 8 randomly chosen discovered attributes. 
The user was asked whether these two set of images represent a consistent visual concept (hence meaningful).
The users were university staffs and students with different knowledge background from various major fields including IT, Electronic Engineering, History, Philosophy, Religion and Classics and Chemical Engineering.
The responses were averaged by considering 1 as meaningful and 0 as non-meaningful.

Table~\ref{tab:noise} illustrates the results of the proposed metric, $\tilde{\gamma}$ compared with the human study. 
In addition, we compare our metric against the MPPCA metric.

\begin{table}[htbp]
  \centering
  \small 
  \caption{
  The results (in percentage) of meaningfulness metric $\tilde{\gamma}$ on each dataset compared with user study and MPPCA metric on ApAy \& ASUN datasets. Each method is configured to discover 32 attributes.
   In addition, for convenience we also report the proposed metric results on AwA and USAA datasets.
    The bold text indicates the top performing method in the proposed metric.
    The higher the more meaningful.
    }
    \scalebox{0.8}{

    \begin{tabular}{c|ccc|ccc|cc|cc}
    \toprule
    \parbox[t]{1.2cm}{Methods\\\textbackslash Datasets} & \multicolumn{3}{c|}{ApAy} & \multicolumn{3}{c|}{ASUN} & \multicolumn{2}{c|}{AwA} & \multicolumn{2}{c}{USAA} \\
    \midrule
          & $\tilde{\gamma}$  & MPPCA  & Human & $\tilde{\gamma}$ & MPPCA & Human & $\tilde{\gamma}$ & Human & $\tilde{\gamma}$ & Human\\
    \cline{2-11}     
    LSH     & 1.7  	&0		& 0    	 & 3.4   & 0		& 0    		& 5.6 		    & \multirow{6}{*}{N/A}  &4.7 & \multirow{6}{*}{N/A} \\
    ITQ     & 4.5  	&34.4	& 20   	 & 16.4  & 31.3		& 22   		& 41.6 		    & 					    &13.1 &\\
    SPH     & 11.7 	&21.9	& 34   	 & 23.8  & 21.9		& 25   		& \textbf{80.7} & 						&\textbf{48.3} &\\
    DBC   	& 8.4  	&15.6	& 32   	 & 24.6  & 21.9		& 30   		& 57.1  		& 						&22.3 &\\
    KSH   	& 38.9 	&37.5	& 60	 & 23.0  & 12.5		& 25   		& 47.6		    &						&18.3 &\\
    PiCoDeS & \textbf{63.3} &\textbf{56.3} &\textbf{71.0} & \textbf{71.5} & \textbf{78.1} & \textbf{43} & 70.5 & & 36.8 &\\
    \bottomrule
    \end{tabular}%
	
  \label{tab:noise}%
  }
\end{table}%
Again, the attribute set discovered by LSH has the
lowest meaningful content at close to 0\%. 
Thus, LSH generates the
least meaningful attribute sets. 
PiCoDeS and SPH generally
discover more meaningful attribute sets. 
The methods using randomization scheme such as LSH and ITQ tend to generate lesser meaningful attribute sets with attribute meaningfulness around 1\%-20\%. 
The results indicate that the attribute meaningfulness could be significantly increased (\ie on average by 10-20 percentage points) by applying learning techniques such as PiCodes, DBC and SPH. 

Compared with the results of the proposed metric $\tilde{\gamma}$, similar trends have been observed in the user study.
Moreover, the user study results compared with $\gamma_{\operatorname{cvx}}$ and $\gamma_{\operatorname{jp}}$ are also shown in Fig.~\ref{fig:calibration}. 
Consistent similar trend as shown in previous experiments is visible.

As for MPPCA metric, similar results can be found such as LSH discovers the least meaningful attribute sets and PiCoDeS generally discover more meaningful attribute sets. 
However, result of our proposed method is closer to the human study in terms of the ranking order of attribute discovery methods. 
This could indicate that the amount of AMT attributes used to train the MPPCA may not be sufficient. 
We note that, the MPPCA was originally designed to have human feedback in multiple iterative process to discover attributes~\cite{parikh2011interactively}. 
As, in our experiment we only fed the MPPCA once with the AMT attributes, it may lack of human feedback.

\begin{figure}[htb]
    \centering
    \includegraphics[width=1\linewidth]{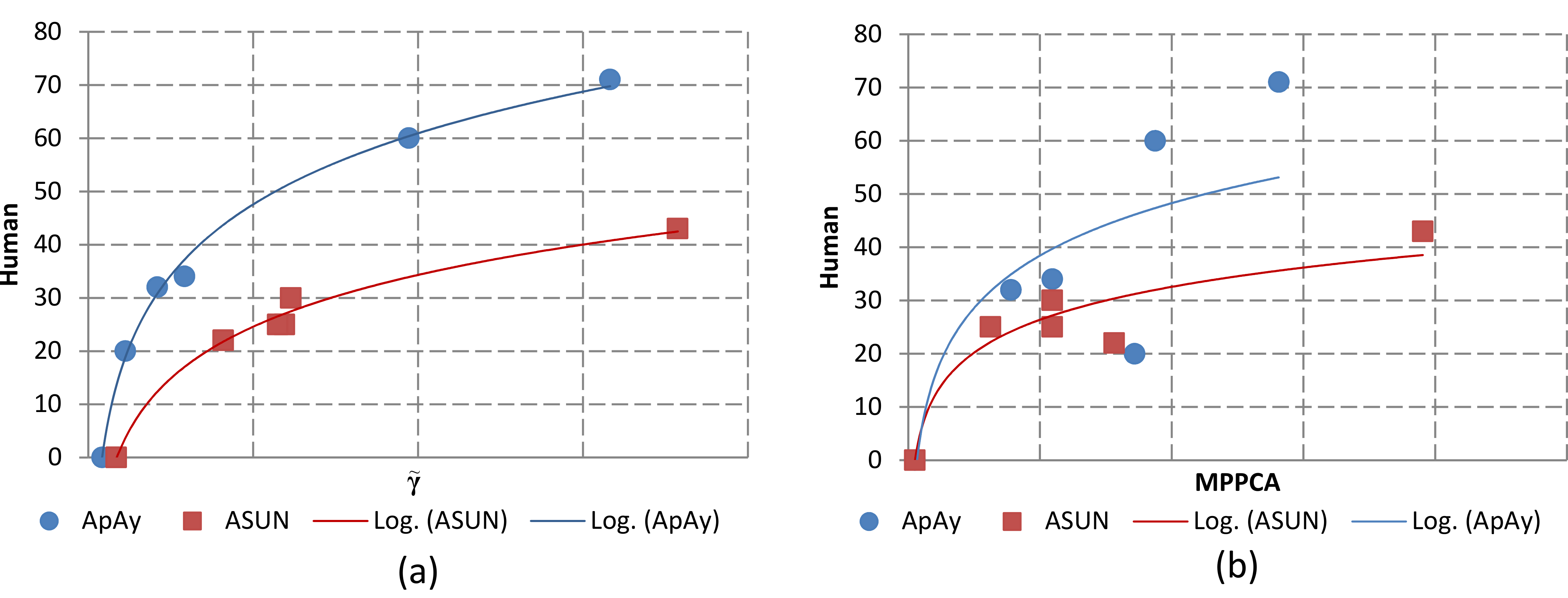}
        \caption{Demonstration of correlation analysis between user study and the proposed method $\tilde{\gamma}$ as well as MPPCA on both ApAy and Asun datasets.}
             \label{fig:calibration_plot}
\end{figure}

We also perform two statistical analysis to compare which metric is closer to the human study. 
Both of the analysis are shown in Fig.~\ref{fig:calibration_plot} by applying a simple logarithmic fitting using the data from Table~\ref{tab:noise}.
Fig.~\ref{fig:calibration_plot}(a) shows that regression line fits these data very well.

The coefficients of determination $R^2$~\cite{draper2014applied} that indicates how well data fit a statistical model for ApAy dataset and ASUN dataset fitting are respectively 0.99 and 0.98 between the proposed metric and the human study results.
The results suggest the regression line of our proposed method nearly perfectly fits the data.
Fig.~\ref{fig:calibration_plot}(b) shows the fitting result for MPPCA is not as good as the proposed method.
The $R^2$ values of MPPCA metric are respectively 0.64 and 0.89 on ApAy dataset and ASUN dataset.

This demonstration further indicates to some extent, our proposed metric is able to evaluate the meaningfulness of a set of discovered attributes from comparative methods as human does.

%

It is noteworthy to mention that the time cost of the evaluation by our metric is significantly lower than the manual process using AMT.
Recall that, the time required for a human annotator (an AMT worker) to finish one HIT is 2 minutes, an AMT worker may need 320 minutes to finish evaluating  5 methods wherein each is configured to discover 32 attributes.
Our approach only needs 105 seconds in total to evaluate all four datasets (\ie 35 seconds each);
thus, leading to several orders of magnitude speedup!

\section{Conclusions}
\label{sec_conclusion}

In this paper, we studied a novel problem of measuring the meaningfulness of automatically discovered attribute sets.
To that end, we proposed a novel metric, here called the \textit{attribute meaningfulness} metric.
We developed two distance functions for measuring the meaningfulness of a set of attributes.
The distances were then calibrated by using subspace interpolation between Meaningful Subspace and Non-meaningful/Noise Subspace.
We extended our previous work by proposing the meaningful attribute set selection technique that leads to a better meaningful subspace approximation.
The final metric score indicates how much meaningful content is contained within the set of discovered attributes.
In the extensive experiment, the proposed metrics were used to evaluate the \textit{meaningfulness} of attributes discovered by two recent automatic attribute discovery methods and four hashing methods on four datasets.
A user study on two datasets showed that the proposed metric has strong correlation to human responses.
Our metric was also shown to be more correlated with the user study compared with a metric adapted from a recent semi-supervised attribute discovery method.
All results suggested that there is a strong indication that the shared structure may exist among the meaningful attributes.
The results also suggest that discovering attributes by optimising the attribute descriptor discrimination and/or 
preserving the local similarity structure 
could yield more meaningful attributes.
In future work, we plan to explore other constraints or optimisation models~\cite{li2016generalized} to capture the hierarchical structure of semantic concepts.
Up to our knowledge, there are still no such works in deep learning area that have similar purpose as our work.
In future, we plan to get more inspiration from the semantic learning in deep learning area and further develop our work in that direction.
We also plan to perform more large-scale user studies using AMT on other datasets.

\vspace{1.5ex}
\noindent
\textbf{Acknowledgment}: We deeply appreciate Teng Zhang's valuable discussion and efforts in proofreading and revisions. This project is partly funded by Sullivan Nicolaides Pathology and the Australian Research Council (ARC) Linkage Projects Grant LP130100230. Arnold Wiliem is funded by the Advance Queensland Early Career Research Fellowship.

\appendix

\section{}
\label{appendix}

\noindent
\textbf{Proposition 4.1}
 Let $\tilde{\mathcal{S}} = \mathcal{S}^2 \cup \tilde{\mathcal{N}}$; when $\tilde{\mathcal{N}} = \{ \}$, then the distance $\delta^*$ between 
 $\tilde{\mathcal{S}}$ and $\mathcal{S}^1$ (refers to \eqref{eq:set}) is minimized. However, when 
 $\tilde{\mathcal{N}} \rightarrow \mathcal{N}$, then the distance between $\tilde{\mathcal{S}}$ and $\mathcal{S}^1$ is asymptotically close to $\delta^*(\mathcal{N},\mathcal{S}^1; \mathcal{X})$, where $\delta^*$ is the distance function presented previously such as $\delta_{\operatorname{jp}}$ and $\delta_{\operatorname{cvx}}$. More precisely, we can define the relationship as follows:
 \begin{equation}
   \lim_{|\tilde{\mathcal{N}}| \rightarrow \infty} \delta^*(\tilde{\mathcal{S}},\mathcal{S}^1; \mathcal{X}) = \delta^*(\mathcal{N},\mathcal{S}^1; \mathcal{X}).
 \end{equation}

\begin{proof}
Let $\Mat{R}^*$ be the solution for the distance $\delta^*$.
The distance $\delta^*$ can be computed as follows:
\begin{align}
	\delta^*(\mathcal{S}^2 \cup \tilde{\mathcal{N}}, \mathcal{S}^1; \mathcal{X}) & = \frac{1}{|\mathcal{S}^2 \cup \tilde{\mathcal{N}}|} \| \Mat{A} \Mat{R}^* - \Mat{B} \|^2_F \nonumber  \\
	& = \frac{1}{|\mathcal{S}^2 \cup \tilde{\mathcal{N}}|} \sum_{\Vec{b}_i \in \mathcal{S}^2 \cup \tilde{\mathcal{N}}} \| \Mat{A} \Vec{r}_i^* - \Vec{b}_i \|_2^2 \nonumber \\
	& = \frac{1}{|\mathcal{S}^2 \cup \tilde{\mathcal{N}}|} \left\{ \sum_{\Vec{b}_j \in {\mathcal{S}^2}} \| \Mat{A} \Vec{r}_j^* - \Vec{b}_j \|_2^2  + \sum_{\Vec{b}_l \in \tilde{\mathcal{N}}} \| \Mat{A} \Vec{r}_l^* - \Vec{b}_l \|_2^2 \right\} 
	\label{eq:finaleq}
\end{align}

As $\mathcal{S}^2$ is assumed to be meaningful and $\tilde{\mathcal{N}}$ is not, then adding attribute $\Vec{b}_l$, a member of $\tilde{\mathcal{N}}$, should increase the average distance. Thus, we have the following inequality:
\begin{equation}
	\frac{1}{|\mathcal{S}^2 \cup \tilde{\mathcal{N}}|} \left\{ \sum_{\Vec{b}_j \in {\mathcal{S}^2}} \| \Mat{A} \Vec{r}_j^* - \Vec{b}_j \|_2^2  + \sum_{\Vec{b}_l \in \tilde{\mathcal{N}}} \| \Mat{A} \Vec{r}_l^* - \Vec{b}_l \|_2^2 \right\} \geq \frac{1}{|\mathcal{S}^2|} \sum_{\Vec{b}_j \in {\mathcal{S}^2}} \| \Mat{A} \Vec{r}_j^* - \Vec{b}_j \|_2^2 
\end{equation}

It means that the distance between $\tilde{\mathcal{S}} = \mathcal{S}^2 \cup \tilde{\mathcal{N}}$ and $\mathcal{S}^1$ can only be minimized when $\tilde{\mathcal{N}}$ is an empty set.
On the other hand, when we keep increasing the size of $\tilde{\mathcal{N}}$, the contribution of the second term in~\eqref{eq:finaleq} becomes more significant than the first term. Thus,~\eqref{eq:finaleq} is approximately close to:
\begin{align}
 \approx \frac{1}{| \tilde{\mathcal{N}}|} \left\{ \sum_{\Vec{b}_l \in \tilde{\mathcal{N}}} \| \Mat{A} \Vec{r}_l^* - \Vec{b}_l \|_2^2 \right\},\text{ as } |\tilde{\mathcal{N}}| >> |\mathcal{S}^2|
\end{align}
In addition, as $|\tilde{\mathcal{N}}| \rightarrow \infty$, $\tilde{\mathcal{N}}$ will be close to $\mathcal{N}$. Thus, the above equation is approximately close to:
\begin{align}
 \approx \frac{1}{| {\mathcal{N}}|} \left\{ \sum_{\Vec{b}_l \in {\mathcal{N}}} \| \Mat{A} \Vec{r}_l^* - \Vec{b}_l \|_2^2 \right\} \nonumber \\
  = \delta^*(\mathcal{N}, \mathcal{S}^1; \mathcal{X})
\end{align}

\end{proof}



\bibliography{egbib}


%
%

\vspace{8pt}
{\small
\textbf{Liangchen Liu} received his BEng and MSc respectively from Chongqing University in 2009 and 2012.
Currently, he is a PhD student in The University of Queensland (UQ).
His research interests are in the areas of computer vision, machine learning and pattern recognition especially visual attributes and their related applications.

\textbf{Arnold Wiliem} is a research fellow at UQ. He received his PhD in 2010 from Queensland University of Technology. He is a member of the IEEE and served as reviewer in various computer vision venues. His interests include Semantic feature extraction, Biometrics, Digital Pathology, Statistical pattern recognition and Nonlinear manifold analysis.

\textbf{Shaokang Chen} received his BEng from South China University of Technology in 1999 and PhD in Electrical Engineering from The University of Queensland in 2005. He was a research fellow in School of ITEE in University of Queensland (2005 - 2008) and a research scientist in National ICT Australia (2008 - 2011). He is currently a Visiting Scholar of The University of Queensland. He was the reviewer for some main journals and conferences in the field of Computer Vision and Pattern Recognition. His research interests include techniques and applications in digital image processing, real-time video analysis, large scale image and video search,  manifold-based classification, and robust face recognition.

\textbf{Brian C. Lovell} received his PhD in 1991 from UQ. Professor Lovell is Director of the Advanced Surveillance Group at UQ. He was President of the International Association for Pattern Recognition (IAPR) [2008-2010], and is Fellow of the IAPR, Senior Member of the IEEE.
His interests include Biometrics, Nonlinear Manifold Learning, and Pattern Recognition.}

\end{document}